\documentclass[runningheads]{llncs}

\usepackage{eccv}

\usepackage{import}
\usepackage{microtype}
\usepackage{graphicx} 
\usepackage{booktabs} 
\usepackage{caption}
\usepackage{xspace} 
\usepackage{tikz}
\usetikzlibrary{positioning} 
\usetikzlibrary{arrows.meta}
\usepackage{multirow}
\usepackage{pgfplots}

\usepackage{amsmath, amssymb, amsfonts} 

\usepackage{bbm}
\usepackage{bm}

\usepackage[normalem]{ulem} 
\usepackage{pifont}

\usepackage{float}

\usepackage[linesnumbered, algoruled, noend, noline]{algorithm2e}

\usepackage{diagbox}

\usepackage{enumerate}
\usepackage{comment}
\usepackage{tikz}
\usetikzlibrary{shapes.geometric}
\subimport{./}{macros}

\usepackage{eccvabbrv}

\usepackage[accsupp]{axessibility}  

\usepackage[breaklinks,colorlinks]{hyperref}

\hypersetup{
    colorlinks=true, 
    linkcolor=blue,  
    citecolor=eccvblue,  
    filecolor=magenta, 
    urlcolor=blue    
}

\usepackage{orcidlink}

\pgfplotsset{compat=1.18}
\begin{document}

\title{SpaceJAM: a Lightweight and Regularization-free Method for Fast Joint Alignment of Images} 

\titlerunning{SpaceJAM: A Method for Fast Joint Alignment of Images}

\author{
Nir Barel* \orcidlink{0000-0003-4751-5653} \and
Ron Shapira Weber* \orcidlink{0000-0003-4579-0678} \and
Nir Mualem\orcidlink{0009-0005-1216-6774} \and
Shahaf E. Finder\orcidlink{0000-0003-0254-1380} \and
Oren Freifeld\orcidlink{0000-0001-9816-9709}
}

\authorrunning{N.~Barel \etal}

\institute{The Department of Computer Science, Ben-Gurion University
of the Negev, 
Israel\\
\email{\{banir,ronsha,nirmu,finders\}@post.bgu.ac.il, orenfr@cs.bgu.ac.il}}

\maketitle

\begin{abstract}

The unsupervised task of Joint Alignment (JA) of images is beset by challenges such as high complexity, geometric distortions, and convergence to poor local or even global optima. Although Vision Transformers (ViT) have recently provided valuable features for JA, they fall short of fully addressing these issues. Consequently, researchers frequently depend on expensive models and numerous regularization terms, resulting in long training times and challenging hyperparameter tuning. We introduce the Spatial Joint Alignment Model (SpaceJAM), a novel approach that addresses the JA task with efficiency and simplicity. SpaceJAM leverages a compact architecture with only $\sim$16K trainable parameters and uniquely operates without the need for regularization or atlas maintenance. Evaluations on SPair-71K and CUB datasets demonstrate that SpaceJAM matches the alignment capabilities of existing methods while significantly reducing computational demands and achieving at least a 10x speedup. SpaceJAM sets a new standard for rapid and effective image alignment, making the process more accessible and efficient. Our code is available at: \url{https://bgu-cs-vil.github.io/SpaceJAM/}.
  
\end{abstract}

\newcommand\blfootnote[1]{
    \begingroup
    \renewcommand\thefootnote{}\footnote{#1}
    \addtocounter{footnote}{-1}
    \endgroup
}

\blfootnote{*Contributed equally.}

\section{Introduction}\label{Sec:Intro}

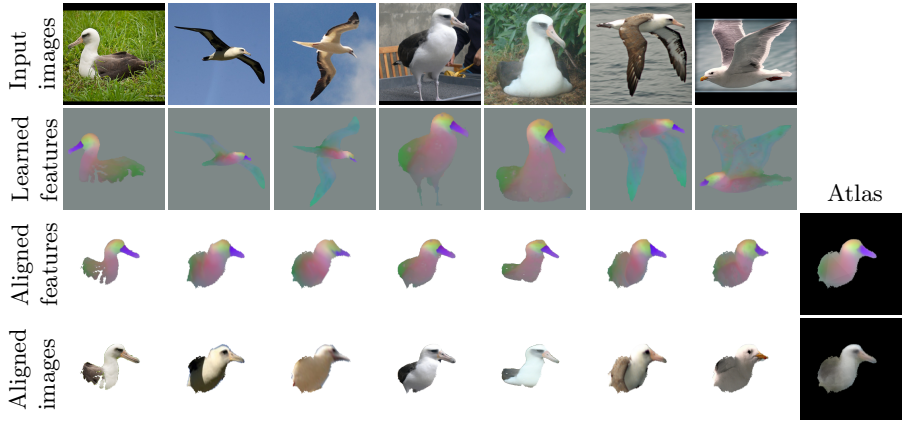
\begin{figure}[ht]
\centering 

\begin{tikzpicture}
\def\numrows{4}
\def\numcols{8}

\def\rowcaptions{{"Input\\images", "Learned\\features", "Aligned\\features", "Aligned\\images", "Encoded\\overlay"}}
\def\directories{{"img", "encoded", "encoded_warped", "img_warped", "img_overlay"}}
\def\imagenames{{"01", "03","04","05","06","07","11","99"}}

\foreach \row in {1,...,\numrows}{
    \pgfmathsetmacro\ypos{-(\row - 1) * 1.4 - 0.75}
    
    \node[rotate=90, align=left] at (-0.41, \ypos) {\pgfmathparse{\rowcaptions[\row-1]}\pgfmathresult};

    \pgfmathparse{\directories[\row-1]}
    \edef\currentdir{\pgfmathresult}
    \xdef\currentdir{./figures/panels_fig_intro/\currentdir} 
    
    \foreach \col in {1,...,\numcols}{
        \pgfmathsetmacro\xpos{\col * 1.4 - 0.75}
        
        \pgfmathsetmacro\imagename{\imagenames[\col-1]}

        \node[inner sep=0pt] (image-\row-\col) at (\xpos, \ypos) {
            \expandafter\includegraphics\expandafter[width=.11\textwidth]{\currentdir/fig_\imagename.png}
        };
    }
}
    \node[font=\footnotesize] (atlas-text) at (10.5,-2.6) {Atlas};
\end{tikzpicture}
\caption{\textbf{SpaceJAM joint alignment} Our framework jointly aligns a set of images of an object category in only a few minutes. Top-to-bottom: 1) input images; 2) learned low-dimensional representations; 3)  aligned features; 4) aligned images. The last column depicts the average representation (atlas) obtained after training.}
\label{fig:intro}
\end{figure}

Joint alignment (JA) of images is the task of geometrically transforming an image collection in a way that optimizes some criterion of mutual similarity. JA is useful for reducing uninformative intra-dateset (or intra-class) variability, thereby facilitating more accurate analysis and interpretation across various applications, ranging from medical imaging to automated recognition systems.
Unfortunately, achieving a successful JA is fraught with challenges, including poor local optima or even poor \emph{global} optima (\ie, trivial non-useful solutions; \eg, shrinking all images to zero size), intra-class variations in pose/orientation/illumination, complex geometric deformations, visual clutter, and more.  
A JA method typically consists of four components: 
1) features to be aligned; 
2) a criterion of joint similarity; 
3) an optimization process;  
4) warping the images via either a parametric transformation family (\eg, homographies) or dense pixel-to-pixel correspondences. 
 
 Traditionally, JA has relied on classical methods such as congealing~\cite{Miller:CVPR:2000:learning,Learned:PAMI:2006:align,Cox:CVPR:2008:LS,Cox:ICCV:2009:LS}, which gradually aligns one image towards the rest, or alignment to a centroid which uses a reference image or a (possibly-latent) template.
Other methods have utilized feature-based approaches, such as SIFT~\cite{Lowe:ICCV:1999:SIFT}, to establish correspondences between images or key-points. While effective in certain domains (\eg, medical scans), those classical approaches often struggle with more diverse image collections.

With the integration of Deep Learning (DL) into JA (\eg, Chelly \etal~\cite{Chelly:CVPR:2020:JA-POLS}), especially via the use of Vision Transformers (ViT)~\cite{Dosovitskiy:2020:ViT} features, the field has witnessed a leap forward. ViTs, DINO specifically~\cite{Caron:ICCV:2021:VITDINO}, offer rich semantic features that partially mitigate some of the challenges that hindered traditional JA methods. However, many issues persist even when using ViT features. This is arguably what led recent works~\cite{Ofri:CVPR:2023:neuracongealingl,Gupta:ICCV:2023:ASIC} to over-rely on high-capacity-but-expensive models,  
as well as extensive regularization strategies. The latter, besides adding to the computational burden, require regularization hyperparameter tuning. Together, that approach led to methods that are slow and often too brittle. 

We take a different approach. While we are happy to use DINO features, \emph{we argue that feeding such features
to high-capacity models is not, in fact, needed for the JA task and that rather than using such models here, it is
better to explore what the best ways are to formulate, and then  solve, this task}. 
With this in mind, this paper shows how to effectively, and efficiently, solve the JA task.
To that end, 
we introduce the Spatial Joint Alignment Model (SpaceJAM), a regularization-free JA approach that 
does not require explicitly maintaining an atlas. See also~\autoref{fig:intro}. SpaceJAM utilizes a lightweight recurrent Spatial Transformer Network (STN)~\cite{Jaderberg:NIPS:2015:STN}, a Lie-algebraic framework, and a novel loss function. 
The proposed approach not only ensures fast JA (\eg, at least a 10x speedup in comparison to contemporary methods~\cite{Ofri:CVPR:2023:neuracongealingl,Gupta:ICCV:2023:ASIC}) but also maintains the geometric integrity of the aligned images. 
We evaluate our method on the SPair-71K and CUB datasets, and show performance either better than, or on par with, state-of-the-art models but with significantly reduced computational demands, marking a significant advancement in the field of JA.

To summarize, our key contributions are as follows:
1)  we introduce SpaceJAM, a lightweight method for JA of in-the-wild images which is much faster than contemporary approaches; 
  2) a novel inverse-compositional loss function that obviates 
    the need of using regularization terms and does not require maintaining an atlas;
    3) an efficient solution for handling reflections (also known as flips).

\section{Related Work}\label{Sec:Related}
\subsubsection{Classical JA methods} focused on techniques that leverage geometric transformations and hand-crafted features. Congealing, a seminal concept introduced in~\cite{Miller:CVPR:2000:learning,Learned:PAMI:2006:align},  refers to gradually aligning an image set by iteratively aligning one image towards the others, to minimize a global cost function, typically the entropy or least-squares of the raw pixel values~\cite{Cox:CVPR:2008:LS,Cox:ICCV:2009:LS,Learned:PAMI:2006:align} or SIFT descriptors~\cite{Huang:CVPR:2007:unsupervised,Lin:CVPR:2012:aligning,Shokrollahi:CVPR:2015:aligngraph}. Other approaches use clustering to simultaneously partition the data and align all class members to their respective mean~\cite{Liu:ICCV:2009:simultaneous,Mattar:2012:unsupervised} or via template matching~\cite{Jain:TPAMI:1996:objectTemplate,Gavrila:ICPR:1998:multi,Felzenszwalb:CVPR:2007:hierarchical}. The success of those classical JA methods, however, was hindered by the quality of the to-be-aligned features. 
Another classical approach seeks to model an image set as a low-rank linear subspace\cite{Kemelmacher:CVPR:2012:collectionflow, He:IVC:2014:GRASTA,Zhang:TPAMI:2019:robust}, the most noticeable work being RASL~\cite{Peng:TPAMIL2012:RASL}, whose main problem lies in its reliance on a good initial alignment.

\subsubsection{Deep learning. }
The advent of deep learning has significantly expanded the capabilities of image alignment techniques. Huang \etal applied the classical congealing algorithm to the features of a Convolutional Neural Net (CNN)~\cite{Huang:NIPS:2012:learning}. The Spatial Transformer Net (STN)~\cite{Jaderberg:NIPS:2015:STN} introduced a differentiable module that can be integrated into larger neural nets to predict and apply spatial transformations to input images or feature maps, enabling end-to-end learning of the alignment process. Since their inception, STNs have proven an invaluable tool for the JA task and were used in both the congealing framework~\cite{Annunziata:ICCV:2019:jointly}, explicit Atlas building~\cite{Lin:CPVR:2017:inverse, Dalca:NIPS:2019:learning,Shapira:NIPS:2019:DTAN, Sinclair:MIA:2022:ISTN}, joint clustering and alignment tasks~\cite{Monnier:NIPS:2020:DTIC,Loiseau:ICV:2021:representing}, 
and for a JA step before building moving-camera background models~\cite{Chelly:CVPR:2020:JA-POLS,Erez:2022:ECCV:MCBM}. Coupled with Generative Adversarial Networks (GAN)~\cite{Goodfellow:NIPPS:2014:GAN}, STNs were also shown to produce high-fidelity class-category canonical spaces (or atlases) for natural images~\cite{Mu:CVPR:s022:coordgan,Peebles:CVPR:2022:gangealing}. However, GAN-based methods requires large amount of training data.

\subsubsection{Semantic correspondence through self-supervision. }
Self-supervised learning (SSL) approaches have recently gained traction for the task of correspondence discovery between images. By leveraging the representations learned by ViTs (such as DINO~\cite{Caron:ICCV:2021:VITDINO}) or text-to-image diffusion models~\cite{Saharia:NIPS:2022:Diffusion}, several works were able to use these features 
for the task of \textbf{pairwise} image correspondence and alignment~\cite{Jeon:ECCV:2018:parn,Jeon:TPAMI:2021:parnj,Zhang:NIPS:2024:Tale2Features,Mariotti:2023:improving, Tang:NIPS:2024:emergentSD}. While useful in the pairwise case, finding \emph{joint dense correspondence} between $N$ images is intractable for a large $N$, as the complexity is $O(N^2)$. Additionally, the Nearest Neighbor (NN) algorithm requires a comparison between every pixel in the source image and all of the pixels in all of the other images. Finally,  methods such as~\cite{Zhang:NIPS:2024:Tale2Features,Tang:NIPS:2024:emergentSD} require multiple $t$ steps for the diffusion model, which further contributes to the computation burden. In contrast, our proposed JA method produces such a mapping without performing NN searches.

\subsubsection{JA using DINO features.}
DINO features offer robust and semantically meaningful representations for many computer-vision tasks, JA included. 
The Neural Congealing algorithm~\cite{Ofri:CVPR:2023:neuracongealingl} utilizes a test-time training framework to build an atlas 
for a given class (\eg, birds) by matching DINO features via rigid and non-rigid warps, using a ResNet-based STN for each~\cite{He:ECCV:2016:resnet}. 
We remark that despite its name, Neural Congealing~\cite{Ofri:CVPR:2023:neuracongealingl} is more related to atlas-based methods than to congealing methods. Such ambivalent terminology is prevalent, unfortunately, as sometimes people mistakenly refer to the JA \emph{task} itself as congealing, regardless whether the JA \emph{method} is congealing-based or atlas-based.

ASIC~\cite{Gupta:ICCV:2023:ASIC} uses DINO features to perform dense warping to map every pixel in the input to a canonical space using a U-net architecture~\cite{Ronneberger:MICCAI:2015:Unet}. 
 However, both~\cite{Ofri:CVPR:2023:neuracongealingl} and~\cite{Gupta:ICCV:2023:ASIC}  grapple with computational overhead, stability issues, and the necessity for extensive regularization to prevent model collapse or trivial solutions. Additionally, ASIC's dense warping approach is prone to incoherent global alignment, often resulting in fragmented or an inconsistent alignment. 
 In contrast, SpaceJAM reaches competitive results much faster and with orders of magnitude fewer trainable parameters. See, \eg,~\autoref{tab:comparison_methods}, as well as the supplementary material (\textbf{SupMat}) for additional running times.
\begin{table}[t]
\newcommand{\cmark}{\ding{51}}%
\newcommand{\xmark}{\ding{55}}%
\centering
\caption{A comparison with recent JA methods (SPair's `Cat' category~\cite{Min:2019:spair}).}
\scriptsize
\label{tab:comparison_methods}
\begin{tabular}{lccccccc}
\toprule
\textbf{Method} & \# Params & \# Losses & \#HP & Atlas-free learning & \#epochs & Time  & PCK@0.10 \\ 
\midrule
 NeuCongealing~\cite{Ofri:CVPR:2023:neuracongealingl} &$28.7\mathrm{M}$ & 8 & 8 & \xmark  & 8K & 1:17:02  &  53.3 \\ 
 \midrule
 ASIC~\cite{Gupta:ICCV:2023:ASIC} &  $7.9\mathrm{M}$& 4 & 5 & \xmark  & 20K  & 1:06:48 & 54.8 \\ 
 \midrule
 SpaceJAM (Ours) & \textbf{0.016M} &  \textbf{1} &  \textbf{0} & \cmark  & \textbf{0.7K} & \textbf{0:05:58}  & \textbf{60.8} \\ 
 \bottomrule
\end{tabular}
\end{table}


\subsubsection{In summary,} 
\begin{figure}[t]
\centering
\def\figwidth{0.99\linewidth}
    \includegraphics[trim = 50mm 0mm 22mm 5mm, clip, width=\figwidth]{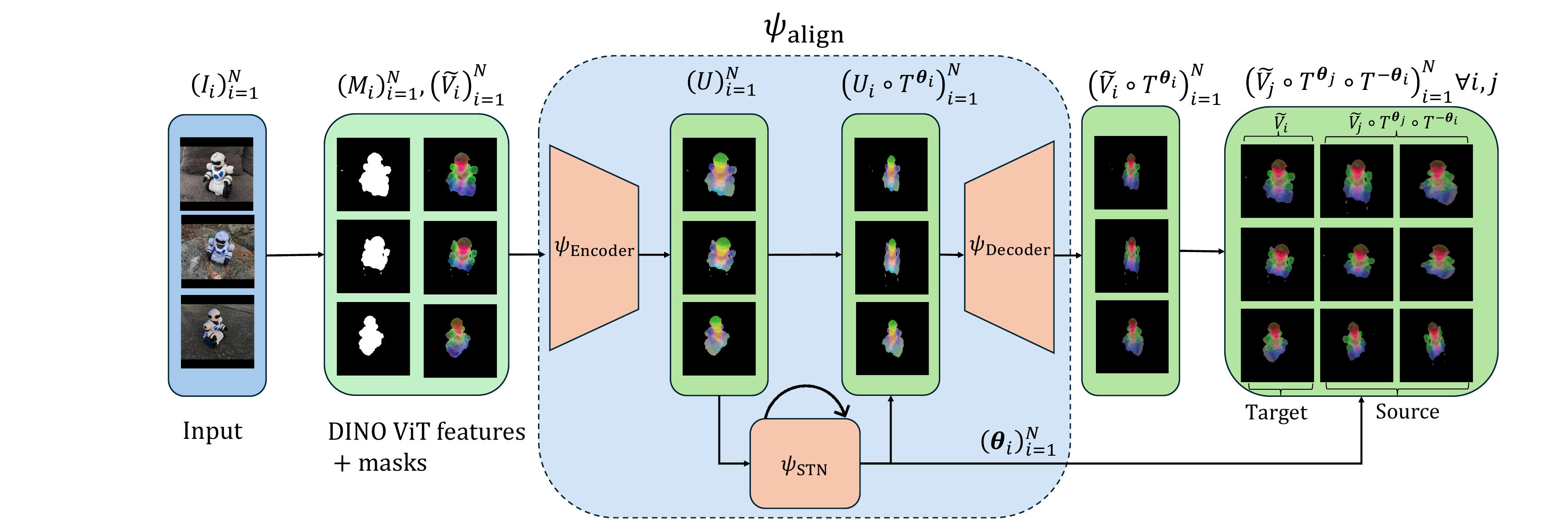}
    \caption{
     \textbf{Framework overview}. Given a set of images,$(I_i)_{i=1}^N$, their DINO-ViT representations and coarse 
     masks 
     , $(V_i, M_i)_{i=1}^N$, SpaceJAM learns an inverse-compositional pairwise alignment between each image pair, and consequently, features, $(U_i)_{i=1}^N$, in a shared semantic space,
     where they are warped (according to learned warping parameters, $(\btheta_i)_{i=1}^N$)
     to produce their aligned versions, $(U_i\circ T^{\btheta_i})_{i=1}^N$. Pairwise alignment of $I_j$ to a target image $I_i$, is achieved by warping to the shared space and then wapring
     the result by the inverse transformation of the target image, yielding $I_j\circ T^{\btheta_j}\circ T^{-\btheta_i}$. 
    }
\label{fig:model:arch}
\end{figure}

while classical methods have provided a strong foundation for image alignment, DL has significantly enhanced the ability to handle complex and varied alignment tasks. The move towards SSL and semantic correspondence methods further illustrates the evolving landscape of the image alignment field, highlighting ongoing challenges and the need for innovative solutions to address computational efficiency, coherence, and optimization stability.
These needs motivated our proposed framework, whose overview is depicted in~\autoref{fig:model:arch}.  
\section{Background: Typical Challenges in Joint Alignment}\label{Sec:Challenge}
Let $\Tcal$ be a family of spatial transformations, from $\Rtwo$ to $\Rtwo$. 
For concreteness, our notation below assumes $\Tcal$ is parametric (as is the case, \eg, for affine transformations or homographies) but most of the discussion holds for the nonparametric case as well. Let $T^\btheta\in\Tcal$ denote 
the generic element of $\Tcal$ where $\btheta$ is the corresponding parameter vector. 
Let $(I_i,I_j)$ be an image pair. 
The \textbf{pairwise alignment} problem is 
 \begin{align}\label{Eq:pairwise:parametric}
     \argmin{T^{\btheta_{ji}}\in\Tcal}D(I_i, I_j\circ T^{\btheta_{ji}})
 \end{align}  
 where $\circ$ is function composition, $T^{\btheta_{ji}}$ is the transformation warping $I_j$ towards $I_i$, and $D$ is a discrepancy measure, based on raw pixels or image features. 
 
 \noindent\textbf{Joint alignment.}
 Given $N$ images, $(I_i)_{i=1}^N$, the underlying notion behind the JA problem (see, \eg,~\cite{Frey:CVPR:1999:estimating}) is that for each observed image $I_i$ there is a latent transformation, $T^{\btheta_i}\in\Tcal$, 
 such that $(I_i\circ T^{\btheta_i})_{i=1}^N$ are essentially different realizations of some shared canonical representation.  
In other words, 
the JA solution is the set of warping parameters, $(\btheta_i)_{i=1}^N$,  that warps the entire image ensemble to some shared representation, usually referred to as an \emph{atlas} (or \emph{prototypical image}). Thus, the JA problem is often formulated as finding a latent atlas, $I_\mu$, together with 
the aforementioned parameters:
\begin{align}\label{eq:JA:noreg}
    \argmin{I_\mu,(T^{\btheta_i})_{i=1}^N\in\Tcal} \sum\nolimits_{i=1}^N 
     D(I_\mu , I_i\circ T^{\btheta_i})\,    .
\end{align}

\subsubsection{JA Approach \#1: Alignment to a Shared Atlas.}\label{challenges:subsec:atlas}
One JA approach relies on building a shared atlas, $I_{\mu}$, that minimizes \autoref{eq:JA:noreg}. The atlas could be a learnable parameter~\cite{Monnier:NIPS:2020:DTIC} or the average of the warped images~\cite{Jain:TPAMI:1996:objectTemplate,Felzenszwalb:CVPR:2007:hierarchical,Peng:TPAMIL2012:RASL,Chelly:CVPR:2020:JA-POLS},
    $I_{\mu}=\tfrac{1}{N}\sum\nolimits_{i=1}^N I_i \circ T^{\btheta_i}$.
A key challenge in that approach is that since the $(\btheta_i)_{i=1}^N$
are unknown, so is $I_{\mu}$ and thus the latter must be   
found during the optimization process together with $(\btheta_i)_{i=1}^N$. Note that the optimization may also be amortized via the training of a neural net~\cite{Erez:2022:ECCV:MCBM}.

Unfortunately, \autoref{eq:JA:noreg} can be undesirably minimized by, \eg, (i) removing all images from the domain of $I_{\mu}$
or (ii) severely shrinking or stretching the images.
Both (i) and (ii) represent poor \emph{global} minima: the nonnegative loss becomes zero, yet the solution is useless. 
Additionally, the process heavily relies on a good initial alignment of the set and is prone to converge to poor local minima, which could be visually seen as a blurred representation of the set. See Erez \etal~\cite{Erez:2022:ECCV:MCBM} for a detailed discussion on problems associated with that approach. 
One possible remedy to the issues above is to use a regularization over the predicted transformations (not to be confused with regularization over the network's weights). That is, the JA problem becomes: 
\begin{align}\label{eq:JA}
    \argmin{I_\mu,(T^{\btheta_i})_{i=1}^N\in\Tcal } \sum\nolimits_{i=1}^N 
     D(I_\mu , I_i\circ T^{\btheta_i})+ \Rcal(T^{\btheta_i};\lambda)\,    ,
\end{align}
where the regularization term,  
 $T^{\btheta_i}\mapsto\Rcal(T^{\btheta_i};\lambda)$ is parameterized by \emph{hyperparameters} (HP), $\lambda$. For instance, one may hope, against hope, that penalizing large deviations from the identity transformation would guide the optimization process to a more plausible atlas. 
However, \emph{there is a major problem here whose severity is often swept under the rug}: the need to determine good values for $\lambda$. 

\subsubsection{JA Approach \#2: Congealing.}
The discussion above leads to another JA approach: congealing~\cite{Miller:CVPR:2000:learning, Learned:PAMI:2006:align}.
The congealing algorithm seeks to minimize some JA criterion (\eg, in its original formulation it was entropy)
by iteratively warping each single image towards the other images but only after those images were already warped using the current estimates of their respective transformations. \emph{One appealing property of congealing is that it defines a notion of central tendency} of the data without having to explicitly build an atlas or to rely on a reference image~\cite{Learned:PAMI:2006:align}.
Congealing avoids some of the pitfalls of aligning to a shared atlas, 
though sometimes regularization is still used. 
It was later shown that least squares congealing (LSC) has several advantages over the entropy-based algorithm in terms of optimization and convergence~\cite{Cox:CVPR:2008:LS}. Finally, the 
clever Inverse-Compositional LSC (IC-LSC) method was introduced to handle outlier images and objects being warped outside of the image domain when applying a single warp to the entire set~\cite{Cox:ICCV:2009:LS}. Although not DL-based, IC-LSC is, in some sense, the closest to our approach.

\begin{figure}[t]
\centering 

\begin{tikzpicture}
\def\panelWidth{1.54cm} 
\def\panelHeight{1.464cm}
\def\startX{-3.761} 
\def\startY{-3.69} 
\def\colDistance{-9.9mm} 
\def\lastColDist{0.5mm}
\def\figwidth{0.8\linewidth}
\def\additionalSpacing{0.12mm}
\node[inner sep=0pt] (img1) at (0,0) {
    \includegraphics[trim = 6cm 13.4cm 14.5cm 10mm, clip, width=\figwidth]{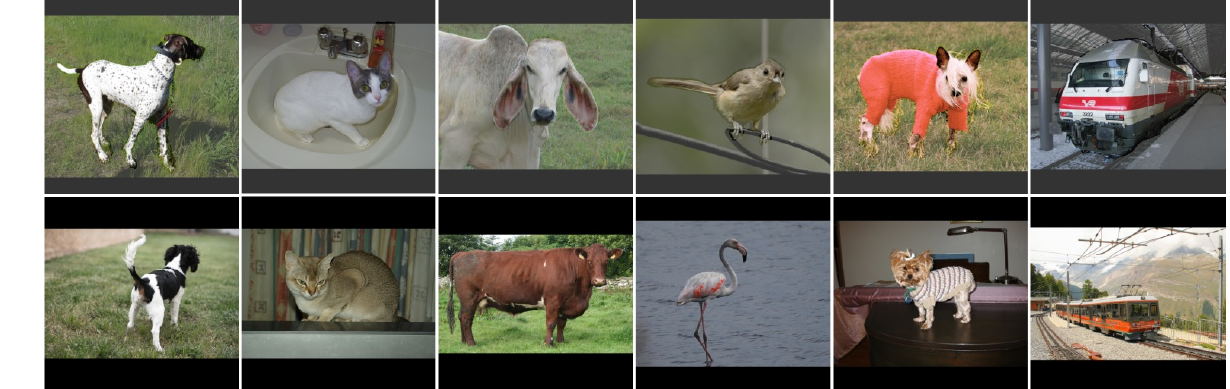}
};
\node[left=of img1, anchor=east, align=center] (source) at (-3.5,0.9) {Source};
\node[left=of img1, anchor=east, align=center] (target) at (-3.5,-0.8) {Target};
\node[inner sep=0pt, below=of img1] (img2) at (0,-0.55) {
    \includegraphics[trim = 6cm 0mm 14.5cm 18cm, clip, width=\figwidth]{../figures/pairwise_comparison.pdf}
};
\node[left=of img2, anchor=east, align=center] (ASIC) at (-3.5,-2.1) {ASIC};
\node[left=of img2, anchor=east, align=center] (ours) at (-3.5,-3.5)  {SpaceJAM\\(Ours)};

\node[inner sep=0pt] (panel1) at (\startX,\startY) 
{\includegraphics[trim = 0mm 0mm 1mm 0mm, clip,width=\panelWidth,height=\panelHeight]{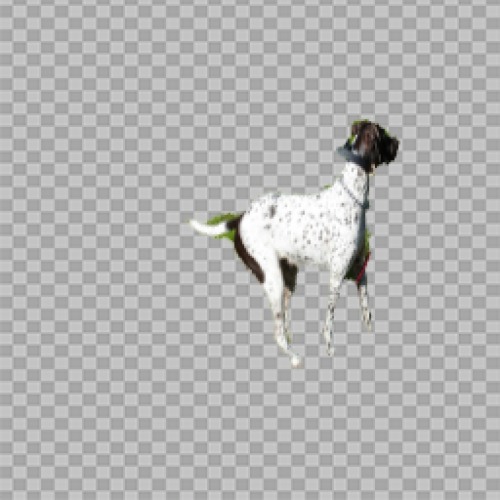}};

\node[inner sep=0pt, right=of panel1, xshift=\colDistance] (panel2) 
{\includegraphics[trim = 0mm 4mm 0mm 4mm, clip,width=\panelWidth,height=\panelHeight]{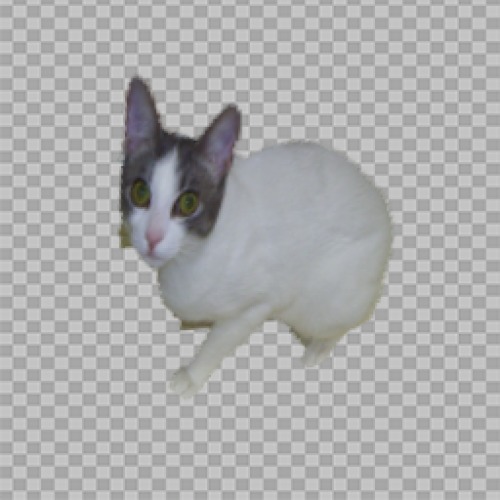}};

\node[inner sep=0pt, right=of panel2, xshift=\colDistance] (panel3)
{\includegraphics[trim = 0mm 0mm 1mm 0mm, clip,width=\panelWidth,height=\panelHeight]{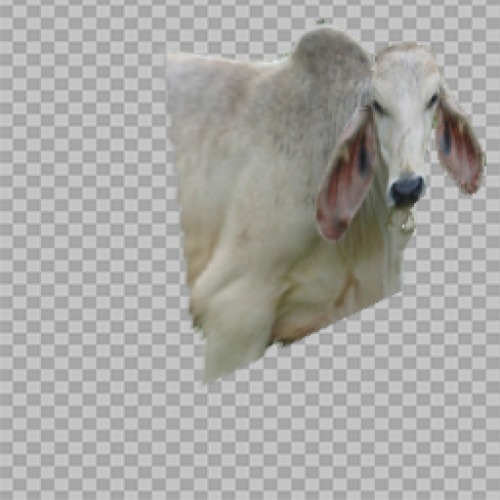}};


\node[inner sep=0pt, right=of panel3, xshift=\colDistance + 0.05mm] (panel4) 
{\includegraphics[trim = 0mm 0mm 1mm 0mm, clip,width=\panelWidth,height=\panelHeight]{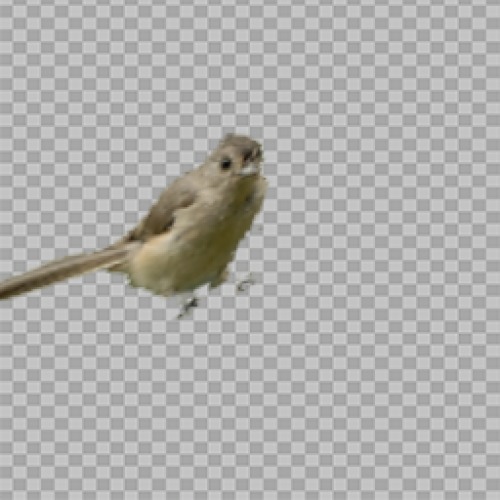}};

\node[inner sep=0pt, right=of panel4, xshift=\colDistance + 0.05mm] (panel5) 
{\includegraphics[trim = 0mm 0mm 1mm 0mm, clip,width=\panelWidth,height=\panelHeight]{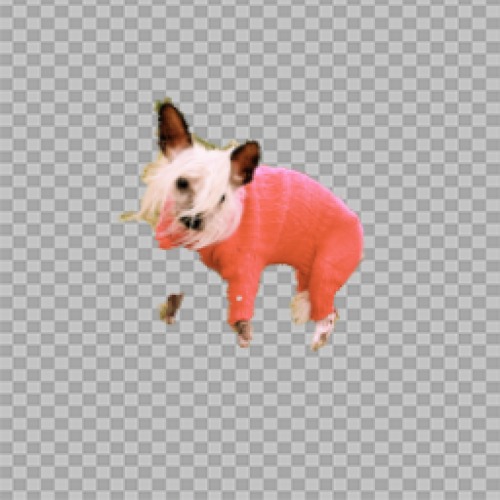}};

\node[inner sep=0pt, right=of panel5, xshift=\colDistance + 0.05mm] (panel6) 
{\includegraphics[trim = 0mm 0mm 1mm 0mm, clip,width=\panelWidth,height=\panelHeight]{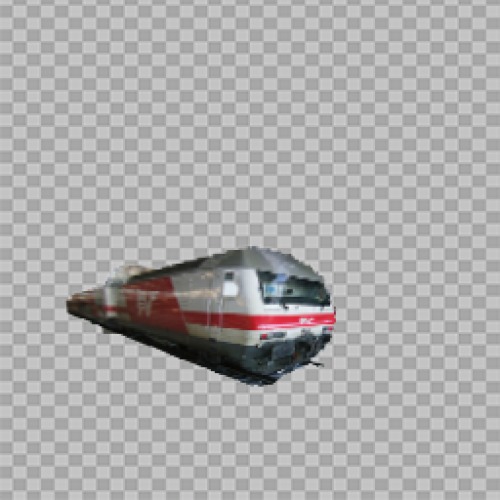}};

\end{tikzpicture}
\caption{\textbf{Global alignment vs. dense warping} evaluated on several classes of the SPair-71K dataset~\cite{Min:2019:spair}. A source image ($1^{\mathrm{st}}$ row) is mapped to the target image
($2^{\mathrm{nd}}$ row) via either dense warping ($3^{\mathrm{rd}}$ row) achieved by ASIC~\cite{Gupta:ICCV:2023:ASIC} (as presented in their paper) or a parametric alignment by SpaceJAM (ours, $4^{\mathrm{th}}$ row). Dense mapping is prone to produce incoherent results (please zoom in to better see that effect) and heavily relies on regularization. In comparison, our  regularization-free method 
produces geometrically-coherent results and is also much faster.}
\label{fig:pairwise:comparison}
\end{figure}

\textbf{Regularization-based DL Approaches for Joint Alignment.}
The aforementioned challenges in solving the JA problem still exist even when considering rich semantically-meaningful representations such as DINO~\cite{Caron:ICCV:2021:VITDINO}. Contemporary methods handle these issues by employing extensive regularization on the predicted warps and/or learned atlas. Consider ASIC~\cite{Gupta:ICCV:2023:ASIC}, for instance, which seeks to densely map every input pixel to a canonical
shared space. Dense mapping strategies are inherently susceptible to breaking the object geometry (as shown in~\autoref{fig:pairwise:comparison}). To mitigate this problem, the authors of~\cite{Gupta:ICCV:2023:ASIC} enforce the following regularization and auxiliary losses: (1) an equivariance loss to handle geometric variations, (2) total-variation regularization to encourage smooth mappings, (3) consistent part alignment to map semantically similar parts, combined with additional two losses that are not directly related to the object geometry, for a total of 5 regularization terms (while~\cite{Ofri:CVPR:2023:neuracongealingl} has 6 reg. terms and 8 HPs). Each loss term requires its own HP, $\lambda$. 
\emph{The reason that the crucial dependency on regularization terms, let alone so many of them, is much worse than one might expect is threefold.} First, the optimal choice of $\lambda$ is usually dataset-specific and this renders such methods either too brittle
or limited. Second, even on a single dataset, determining a good value typically requires some supervision -- at odds with the unsupervised nature of the JA task.
Third, the need to have several (sometimes many) loss terms, can complicate and prolong the optimization process. Thus, it is unsurprising that current DL-based JA methods such as~\cite{Ofri:CVPR:2023:neuracongealingl,Gupta:ICCV:2023:ASIC} must use expensive architectures and 
are slow.  

\section{Method}\label{Sec:Method}
Given a collection of images, $(I_i)_{i=1}^N$, along with their DINO representations, $(V_i)_{i=1}^N$, of an object (\eg, a particular bird) or an object class (\eg, multiple birds), our goal is to jointly align the images in a semantically-meaningful manner.
Before describing the proposed method we emphasize that all of our design choices below were made with the goal of making the JA process easier, faster, and more stable. Together, those choices have a surprisingly strong synergistic effect, yielding a lightweight JA method that is \emph{much} faster and its number of trainable parameters is $<1\%$ of the numbers in contemporary methods. 

\subsection{Preprocessing}\label{method:subsec:preprocessing}
The goal of our preprocessing procedure is to extract  object masks and reduce the dimensionality of the DINO features. 
Similarly to~\cite{Ofri:CVPR:2023:neuracongealingl, Gupta:ICCV:2023:ASIC}, and to keep the comparison with those methods fair,  we use~\cite{Amir:2021:VITdescriptoers} to produce the saliency-based \textbf{masks} for the image collection. Let $M_i$ denote the binary mask for $I_i$. 
Next, we apply  \textbf{dimensionality reduction} to the DINO features. Let $V_i\in \RR^{d\times H \times W}$ where $d$ is typically high (\eg, $d=384$) and $(H, W)$ are the (height,width) of the DINO representation. 
Our dimensionality reduction has two steps. 

To reduce the computational burden, we apply Principal Component Analysis (PCA) using the first $K$ principal components, where $K<d$ (in our experiments, $K=25$).  
Let $\Hat{V_i}\in \RR^{K\times H \times W}$ denote the PCA representation of $V_i$, and let $\Tilde{V}_i=\Hat{V_i}\cdot M_i$ denote its masked version. 

In the second step, we train a fully convolutional, low-capacity autoencoder (only $\sim$3K parameters) to reduce the number of channels to 3. We chose 3 since it is low enough to serve as an information bottleneck that removes redundant details and since it allows for interpretability when visualizing the encoded features using RGB colors. 
The AE is trained to reconstruct $\Tilde{V_i}$ using the following loss, 
\begin{align}\label{eq:recon}
  \Lcal_{\mathrm{AE}} = \sum\nolimits_{i=1}^{N} \ellTwoNorm{ \Tilde{V_i} - \psi_{\mathrm{dec}}(
  \psi_{\mathrm{enc}}(\Tilde{V_i}))}^2\, .
\end{align} 
As we explain later, the obtained latent representation of the AE, $U_i\triangleq\psi_{\mathrm{enc}}(\Tilde{V_i})\in \RR^{3\times H \times W}$, will serve a purpose in our alignment network.
We find that the combination of PCA and AE to reduce the input's number of channels is highly effective from both computational and representational points of view.

\subsection{Alignment Network}
Our alignment network, $\psi_{\mathrm{align}}$, is based on an STN~\cite{Jaderberg:NIPS:2015:STN}.  Below we provide a brief overview of that STN and how we handle common pitfalls in STN training. 
Let $\psi_{\mathrm{STN}}$ denote the STN
and, for now, let $X$ denote its generic input.  
The core of the STN is the so-called \textbf{localization network}, denoted by $\psi_{\mathrm{loc}}$, that predicts transformation parameters from $X$; \ie, $\btheta=\psi_{\mathrm{loc}}(X)$. The output
of the entire STN is $(\btheta, X \circ T^{\btheta})=\psi_{\mathrm{STN}}(X)$. That is, the output is both $\btheta$ and the warped image obtained by applying $T^\btheta$ to $X$. 
STNs have been shown to be an effective tool for various tasks, albeit difficult to optimize~\cite{Lin:CPVR:2017:inverse, Skafte:CVPR:2018:DDTN}. 
The issue stems from a few reasons, which we address in a theoretically-grounded manner. 

The first difficulty is in predicting large transformations. 
To address this issue, the Inverse-Compositional STN (IC-STN)~\cite{Lin:CPVR:2017:inverse} was proposed, which, based on the classical Inverse-Compositional Lucas \& Kanade algorithm~\cite{Lucas:IJCAI:1981:iterative}, predicts a cascade of smaller warps and composes them. 
In our case, we do so by recursively feeding the STN several times with its own output (see \textbf{SupMat} for details).  
Empirically, 5 recurrences suffice 
(a higher number  yielded a very small gain) so this is what we used in our experiments.
Since the recurrences share the same STN, the number of trainable parameters does not increase with the number of recurrences.  
As we apply several transformations in cascade, each predicted transformation can be relatively small. 
A fine nuance is that while usually each STN call involves interpolation (due to the resulting irregular grid), here whenever we compose transformations we perform only the last interpolation; this improves the quality of the resulting image as fewer interpolation artifacts occur. 
Our entire model, $\psi_{\mathrm{align}}$, consists
of the AE and 5 recurrences of $\psi_{\mathrm{STN}}$.

\subsection{Lie Algebras, Lie groups, and the Matrix Exponential}\label{method:subsec:lie:algebra}

Another difficulty is that, without special care, an STN might predict non-invertibile transformation matrices, which could hinder the optimization process; \eg, \cite{Skafte:CVPR:2018:DDTN} showed that restricting an affine STN to invertible affine transformations allows for more robust and stable training. 
Thus, our alignment network, $\psi_{\mathrm{align}}$,
uses a cascade of transformations that are members of a Lie group. 
This formulation is  more stable and robust compared to the vanilla STN. 
We represent spatial transformations as elements of Lie groups via a Lie-algebraic parameterization, as we now explain. 
Consider these 6 spaces of 3-by-3 real matrices: 
\begin{align*}
&\mathrm{se}(2)= \left\{
\bA: \bA=
\left[
\begin{smallmatrix}
0 & \theta_2 & \theta_3 \\ -\theta_2 & 0 & \theta_6 \\ 0 & 0 & 0
\end{smallmatrix}
\right]
\right\}\,,
 \,
\mathrm{SE}(2)= \left\{\bT: \bT=
\left[\begin{smallmatrix} T_1 & T_2 & T_3 \\ T_4 & T_5 & T_6 \\ 0 & 0 & 1 
\end{smallmatrix}\right]
\& \left[\begin{smallmatrix} T_1 & T_2  \\ T_4 & T_5  
\end{smallmatrix}\right]\in \mathrm{SO(2)}
\right\}, \nonumber
\\
&\mathrm{aff}(2)= \left\{\bA: \bA=
\left[\begin{smallmatrix} \theta_1 & \theta_2 & \theta_3 \\ \theta_4 & \theta_5 & \theta_6 \\ 0 & 0 & 0 
\end{smallmatrix}\right]\right\}\,,
\,
\mathrm{Aff}(2)= \left\{\bT: \bT=
\left[\begin{smallmatrix} T_1 & T_2 & T_3 \\ T_4 & T_5 & T_6 \\ 0 & 0 & 1 
\end{smallmatrix}\right]
\& \det\bT>0 \right\}, \nonumber
\\
&
\mathrm{sl}(3)= \left\{\bA: \bA=
\left[\begin{smallmatrix} \theta_1 & \theta_2 & \theta_3 \\ \theta_4 & \theta_5 & \theta_6 \\ \theta_7 & \theta_8 & -(\theta_1+\theta_5) 
\end{smallmatrix}\right]\right\}\,,
\,
\mathrm{SL}(3)= \left\{\bH: \bH=
\left[\begin{smallmatrix} H_1 & H_2 & H_3 \\ H_4 & H_5 & H_6 \\ H_7 & H_8 & H_9 
\end{smallmatrix}\right]
\& \det\bH=1 \right\} .
\end{align*}
 The shortcuts  $\mathrm{SE}$ (or $\mathrm{se}$)
 $\mathrm{Aff}$ (or $\mathrm{aff}$), 
  $\mathrm{SL}$ (or $\mathrm{sl}$), 
  and $\mathrm{SO}$ stand for ``Special Euclidean'', ``Affine'', ``Special Linear'', and ``Special Orthogonal", 
respectively. 
The following are well-known mathematical facts, widely used in computer vision  (see, \eg,~\cite{Lin:CVPR:2009:Lie,Lin:CVPR:2010:Persistent, Chelly:CVPR:2020:JA-POLS,Erez:2022:ECCV:MCBM}): 
1)
    $\mathrm{se}(2)$, $\mathrm{aff}(2)$, and $\mathrm{sl}(3)$ 
      are \emph{linear} spaces, called (matrix) \textbf{Lie algebras}. 
    2)
     $\mathrm{SE}(2)$, $\mathrm{Aff}(2)$, and $\mathrm{SL}(3)$ 
    are \emph{nonlinear} spaces. Moreover, they are (matrix) \textbf{Lie groups}; namely, each of these spaces is both a group (the binary operator being matrix multiplication) and a smooth manifold, and the group operations (multiplication; inversion) are smooth. 
    3)
    Each element of each of those groups is an \textbf{invertible matrix}. 
    4)
    \textbf{Nesting property:} $\mathrm{se}(2)$ is a linear subspace of $\mathrm{aff}(2)$ and the latter 
        is a linear subspace of $\mathrm{sl}(3)$. Likewise,  $\mathrm{SE}(2)$ is a matrix subgroup of $\mathrm{Aff}(2)$ and the latter 
        is a matrix subgroup of $\mathrm{SL}(3)$.
    5)
    $\mathrm{SE}(2)$ is the group of rigid-body transformations in $\Rtwo$. 
      $\mathrm{Aff}(3)$ is the group of (orientation-preserving) \textbf{invertible affine transformations} in $\Rtwo$.
     $\mathrm{SL}(3)$, the group of volume-preserving linear transformations in $\Rthree$,
         can also be identified with the group of \textbf{homographies} in $\Rtwo$ (this is why we denoted its generic element by $\bH$). 
    6)
    The \textbf{matrix exponential} maps $\mathrm{se}(2) \to \mathrm{SE}(2)$,
    $\mathrm{aff}(2)\to\mathrm{Aff}(2)$, and $\mathrm{sl}(3)\to\mathrm{SL}(3)$.  
   

The matrix exponential provides a differentiable map from a linear space (the algebra) into its corresponding nonlinear space (the group).
This implies that, using the Lie-algebraic parametrization, an STN~\cite{Jaderberg:NIPS:2015:STN} can be easily restricted to yield only elements of the Lie group of interest~\cite{Skafte:CVPR:2018:DDTN}. 
This is useful since while optimization over nonlinear manifolds can be hard (\eg, even a simple gradient-descent step might bring us outside the manifold), 
the Lie-algebraic parametrization, together with the matrix exponential, lets us perform gradient-based optimization in a linear space (for other methods for optimization 
on manifolds, see Boumal's textbook~\cite{Boumal:Book:2023:OptMani}) 
In fact, \cite{Skafte:CVPR:2018:DDTN} showed that an STN restricted to \emph{invertible} affine transformations was more stable and yielded better results than the commonly-used unrestricted STN.  
In our case, there is also an additional reason why we use Lie groups and this is the fact
that our approach relies directly on Lie groups' closure under inversion and composition (as detailed below).

\subsection{Loss Function}\label{method:subsec:loss} 
  
Recall that, for image $I_i$, 
we let $V_i, U_i$, and $M_i$ denote its DINO representation, an encoded 3-channel representation, and mask map, respectively. 
We train $\psi_{\mathrm{align}}$ to find the JA of $(I_i)_{i=1}^N$ in a forward-inverse compositional manner. Part of the motivation behind our loss is to avoid the drawbacks of explicitly warping the images to a shared space (see~\autoref{challenges:subsec:atlas}). In particular, we avoid using regularization terms over the transformations. 
Our loss consists of two conceptual steps: (i) for each image, $I_i$, in batch $k$ -- denoted by  $B_{k}=
(I_i)_{i=1}^{N_b}$ (where $N_b$ is the \# of images in the batch) -- predict the forward warping parameters, $\btheta_i$, using $U_i$ as the recurrent STN's input,  
and then (ii) compose the forward warp with the inverse warp of every other image in the batch, to build the following loss: 
\begin{align}\label{eq:loss:full}
   \Lcal_\mathrm{IC} = \sum\nolimits_{i=1}^{N_b}
   \sum\nolimits_{j:j\neq i} \ellTwoNorm{\Tilde{V}_j - (\Tilde{V}_i \circ T^{\btheta_i}
   \circ
   T^{-\btheta{_j}})  
   }^2 \,.
\end{align}
Minimizing this loss implicitly maps each image to the shared space, since 
\begin{align}\label{eq:forward:inverse}
   \Tilde{V_j}  \approx
   \Tilde{V_i}  \circ T^{\btheta_i}
   \circ
   T^{-\btheta{_j}}
 \Leftrightarrow 
   \Tilde{V_j} \circ T^{\btheta{_j}}\approx \Tilde{V_i}  \circ T^{\btheta_i}  
  \, .
\end{align}
The loss in~\autoref{eq:loss:full} relies on Lie groups' closure under inversion (\ie, the matrix associated with $T^{\btheta}$ is invertible) and composition. 
Notably, the proposed loss effectively frees us from having to use
 regularization. This is because, by construction, a ``bad'' predicted transformation would directly increase the loss as the corresponding warped image would fail to ``explain away" the other images, unless all the predicted transformations would be ``bad'' in \emph{exactly} the same way (which is unlikely and indeed never happens in practice).

Training is done via standard DL gradient-based optimization, except we also use a Lie-algebraic curriculum learning to ease the optimization further. Meaning,  
 exploiting the nested structure of the Lie algebras, 
we start the training with $\mathrm{SE}(2)$ and end with homographies. See \textbf{SupMat} for details. After training, SpaceJAM jointly aligns the entire image collection using its forward pass.

\textbf{Reflections.}
The outline of our proposed solution for handling reflections (\ie, flips), whose full details are in our \textbf{SupMat},
is as follows. 
During training, we dynamically change the selected flip configuration and train on all configurations simultaneously (this yields smooth and robust training)
but save computations and increase stability by  \emph{calculating gradients only for the best flip configuration}. 

\textbf{Atlas Building.}
Upon learning the JA, the method can also quickly generate an atlas as follows.  
Given the images $(I_i)_{i=1}^N$,  we predict the forward warping parameters via SpaceJAM's forward pass and obtain 
the aligned images via $(I_i\circ T^{\btheta_i})_{i=1}^N$. The atlas can be taken as the sample mean of either the warped features,  
    $ \frac{1}{N}\sum\nolimits_{i=1}^N V_i \circ T^{\btheta_i}$, 
or the warped PCA-related versions, 
 $ \frac{1}{N}\sum\nolimits_{i=1}^N \Tilde{V}_i \circ T^{\btheta_i}$,
Either way results in a semantically-meaningful atlas; see,~\autoref{fig:congealing}.
\subsection{Implementation}
We use simple architectures. 
 $\psi_{\mathrm{loc}}$ is a 2-layer CNN with $\sim$13K trainable parameters and $\psi_{\mathrm{AE}}$ is a fully-convolutional AE with $\sim$3K trainable parameters, so the total is only 16K. 
The final layer predicts 8 parameters (the number of degrees of freedom in a homography) in total. We predict a cascade of 5 warps in total, where $\psi_{\mathrm{align}}$ is shared throughout the process (akin to standard IC-STN~\cite{Lin:CPVR:2017:inverse}). SpaceJAM is implemented in \texttt{PyTorch}~\cite{Paszke:NISP:2019:pytorch} and optimized for 300 epochs for $\psi_{\mathrm{AE}}$ and then an additional 400 for the entire framework ($\psi_{\mathrm{AE}}$+$\psi_{\mathrm{align}}$), using the Adam optimizer~\cite{Kingma:arxiv:2014:Adam}
and a StepLR scheduler ($\gamma=0.9$, every 50 epochs). 
\section{Results}\label{Sec:Results}
 
\begin{figure}[t]
\centering 

\resizebox{0.95\textwidth}{!}{%
\begin{tikzpicture}
\def\numrows{4}
\def\numcols{8}

\def\rowcaptions{{"Input\\images", "Learned\\features","Pairwise\\alignment","JA\\features", "Aligned\\images"}}
\def\directories{{"img", "overlays","source_to_target", "congealed_enc", "congealed_imgs"}}
\def\imagenames{{"00", "01","03", "04","05","06","07","99"}}

\foreach \row in {1,...,\numrows}{
    \pgfmathsetmacro\ypos{-(\row - 1) * 1.4}
    
    \node[rotate=90, align=left] at (-0.41, \ypos) {\pgfmathparse{\rowcaptions[\row-1]}\pgfmathresult};

    \pgfmathparse{\directories[\row-1]}
    \edef\currentdir{\pgfmathresult}
    \xdef\currentdir{./figures/panels_congealing/\currentdir} 
    
    \foreach \col in {1,...,\numcols}{
        \pgfmathsetmacro\xpos{\col * 1.4 - 0.75}
        
        \pgfmathsetmacro\imagename{\imagenames[\col-1]}

        \node[inner sep=0pt] (image-\row-\col) at (\xpos, \ypos) {
            \expandafter\includegraphics\expandafter[width=.12\textwidth]{\currentdir/fig_\imagename.png}
        };
    }
}

    \node[draw=red, 
          thick, 
          minimum size=1.4cm, 
          anchor=south,
          ] (redsquare) at (0.64,-3.55) {}; 

    \node[below=-0.05cm of redsquare, text=red, font=\scriptsize] {Source}; 

    \node[font=\footnotesize] (atlas-text) at (10.4,-3.3) {Atlas};

\end{tikzpicture}
}
\caption{\textbf{Pairwise and joint alignment using SpaceJAM.} The input images ($1^\mathrm{th}$ row) overlayed by the learned features ($2^\mathrm{nd}$ row) which are used to predict the warping parameters. The $3^\mathrm{rd}$ row shows source-to-target alignment under severe conditions and 
the $4^\mathrm{th}$ row shows the jointly aligned features and the category atlas (right column).}
\label{fig:congealing}
\end{figure}
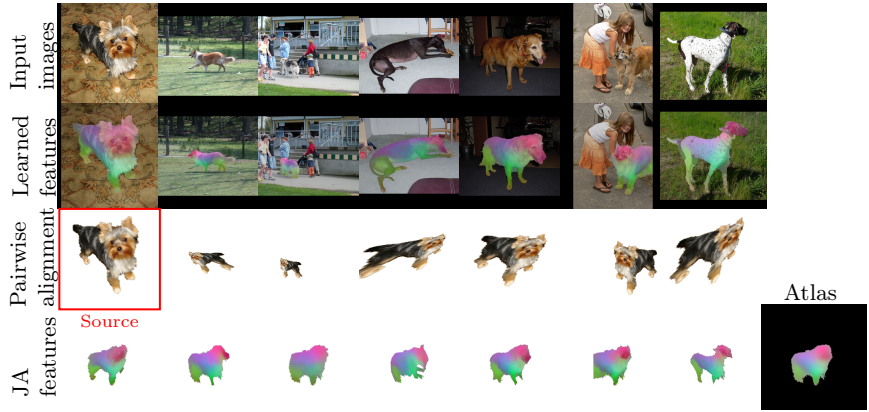

We evaluate our method on several real-world 
benchmarks, containing object categories with different illuminations, visual clutter, and occlusions, and compare it with several JA methods. 
The \textbf{Datasets} include \textbf{SPair-71k}~\cite{Min:2019:spair}, which consists of 1,800 images from 18
categories (\ie, bird). As is common in the weakly-supervised test-time optimization scenario, we train 
(in a weakly-supervised manner; \ie only the categories are known, but the latent alignments are not)
SpaceJAM on each of 
the SPair-71k test sets independently and report results for each category. For \textbf{CUB-200}~\cite{Wah:2011:CUB}, we 1) report results on its first 3 categories, consisting of 25-30 images each when comparing with~\cite{Gupta:ICCV:2023:ASIC} and 2) the average of 14 subsets, following the protocol in~\cite{Ofri:CVPR:2023:neuracongealingl,Peebles:CVPR:2022:gangealing}. For a fair comparison, in the quantitative comparison our method, like~\cite{Ofri:CVPR:2023:neuracongealingl,Gupta:ICCV:2023:ASIC}, used DINOv1 (ViT-S/8)~\cite{Caron:ICCV:2021:VITDINO}. For completeness and future comparisons, we also report our results with DINOv2~\cite{Oquab:2023:dinov2}.

\subsection{Qualitative Results}\label{results:subsec:qualitative}
Recall that SpaceJAM is trained in an inverse-compositional framework which first aligns an image to a shared space and then warps it to all other images in the batch using the inverse transform of the target. This allows us to compute (after the training is done) a shared atlas without explicitly maintaining one during training. 
Examples of our method's JA and resulting atlas can be seen in~\autoref{fig:intro} and ~\autoref{fig:congealing}. \autoref{fig:pairwise:comparison} provides a visual comparison between SpaceJAM and ASIC in terms of pairwise alignment (for AISC, we use the visual results appearing in~\cite{Gupta:ICCV:2023:ASIC}). Evidently, SpaceJAM achieves high fidelity and geometrically-coherent alignment under challenging conditions. 
~\autoref{fig:congealing} shows how the learned low-dimensional features ($U_i\in \RR^{3\times H \times W}$; second row) allow us to perform both pairwise alignment ($3^{\mathrm{rd}}$ row) and JA ($4^{\mathrm{th}}$ row) in an interpretable manner, where semantically-related parts (\eg, faces) share the same color across images.   
The visual results indicate that SpaceJAM can align diverse collections of images efficiently and accurately (see the \textbf{SupMat} for more visual results). 

\subsection{Quantitative Evaluation}\label{results:subsec:PCK}
We assess our method on the semantic point correspondence task using the SPair-71k~\cite{Min:2019:spair} and CUB-200~\cite{Wah:2011:CUB} datasets, which are widely recognized benchmarks in both pairwise and JA domains. The performance is quantified by the PCK-Transfer metric, which calculates the proportion of keypoints accurately aligned within a threshold of $\alpha\cdot\max(h,w)$ relative to the true position. Following established protocols~\cite{Ofri:CVPR:2023:neuracongealingl}, $\alpha=0.1$ (denoted as PCK@0.10) for both datasets, with $(h,w)$ representing the dimensions of the object's bounding box. We perform pairwise alignment using~\autoref{eq:forward:inverse}.

In line with the comparative framework established by~\cite{Gupta:ICCV:2023:ASIC}, we categorize existing methods into 3 groups: (1) \emph{Strong supervision}, which relies on manually annotated keypoints; (2) \emph{GAN supervision}, using category-specific GANs and; (3) \emph{Weak supervision}, using category-level supervision without explicit keypoints. A subset of these methods adopts a train/test paradigm, using a large amount of data. Unlike those works, SpaceJAM involves training a compact model and leverages a test-time optimization scheme. The baseline scores, obtained from~\cite{Gupta:ICCV:2023:ASIC}, include: 1) DINO nearest neighbor (DINO+NN;~\cite{Amir:2021:VITdescriptoers});
2) Neural Congealing~\cite{Ofri:CVPR:2023:neuracongealingl}, which builds a shared atlas; and 3) ASIC~\cite{Gupta:ICCV:2023:ASIC} which densely maps pixels to a canonical grid (both are weakly-supervised JA methods).  

\autoref{tab:spair:full} shows PCK@0.1 for all classes of the SPair-71k dataset~\cite{Min:2019:spair}. We report the average of 3 runs for SpaceJAM. 
Our method performs best on average for all object classes (\textbf{All}) in the \emph{weak supervision} category, using a much lighter model that trains $10\mathrm{x}$ faster. It also performs best and second-best in 9 out of 18 classes. The results are consistent for both rigid objects (\ie, `Train') and ones with extreme variations (`Cat') with the exception of the `Chair' category, where it was unable to overcome the initial coarse masks provided by~\cite{Amir:2021:VITdescriptoers}.~\autoref{tab:CUB} shows the results on the CUB-200 dataset~\cite{Wah:2011:CUB} as reported in~\cite{Gupta:ICCV:2023:ASIC}, where our method is the runner-up after~\cite{Gupta:ICCV:2023:ASIC}. That dataset consists of only birds and offers less variety than SPair-71K. When comparing to~\cite{Ofri:CVPR:2023:neuracongealingl, Peebles:CVPR:2022:gangealing}, on 14 subsets, our method outperforms both. Please see our \textbf{SupMat} for further discussion of the results.

 \begin{table}[t]
\centering
\caption{\textbf{SPair-71k Dataset results:} Per-class and overall mean PCK@$0.10$ evaluated on the test set. The best result among \textit{weakly-supervised} methods is boldfaced (the runner-up is underlined). ($\star$) indicates that the method used a reference image. Missing values ($-$) indicate
unreported results (we, like~\cite{Gupta:ICCV:2023:ASIC}, found that running the code from~\cite{Ofri:CVPR:2023:neuracongealingl}
usually did not converge successfully).}
\label{tab:spair:full}
\resizebox{\textwidth}{!}{
\begin{tabular}{@{}clccccccccccccccccccc@{}}
\toprule
\textbf{Supervision} & 
\textbf{Method} & 
Aero & Bike & Bird & Boat & Bottle & Bus & Car & Cat & Chair & Cow & Dog & Horse & Motor & Person & Plant & Sheep & Train & TV & \textbf{All}\\  

\midrule

\multirow{1}{3.2cm}{\centering Strong Supervision} &
SCorrSAN~\cite{Huang:ECCV:2022:SCorrSAN} & 
57.1 & 40.3 & 78.3 & 38.1 & 51.8 & 57.8 & 47.1 & 67.9 & 25.2 & 71.3 & 63.9 & 49.3 & 45.3 & 49.8 & 48.8 & 40.3 & 77.7 & 69.7 & 55.3\\

\hline

\multirow{1}{3.2cm}{\centering GAN supervision} &
GANgealing~\cite{Peebles:CVPR:2022:gangealing} & 
- & 37.5 & - & - & - & - & - & 67.0 & - & - & 23.1 & - & - & - & - & - & - & 57.9 & -\\

\hline

\multirow{7}{*}{\parbox{3.2cm}{\centering Weak supervision \\ (train/test)}} &
CNNGeo~\cite{Rocco:CVPR:2017:CNNGeo} & 
23.4 & 16.7 & 40.2 & 14.3 & 36.4 & 27.7 & 26.0 & 32.7 & 12.7 & 27.4 & 22.8 & 13.7 & 20.9 & 21.0 & 17.5 & 10.2 & 30.8 & 34.1 & 20.6\\

& A2Net~\cite{Seo:ECCV:2018:A2Net} & 
22.6 & 18.5 & 42.0 & 16.4 & 37.9 & \underline{30.8} & 26.5 & 35.6 & 13.3 & 29.6 & 24.3 & 16.0 & 21.6 & 22.8 & \underline{20.5} & 13.5 & 31.4 & 36.5 & 22.3\\

& WeakAlign~\cite{Rocco:CVPR:2018:weakalign} & 
22.2 & 17.6 & 41.9 & 15.1 & 38.1 & 27.4 & 27.2 & 31.8 & 12.8 & 26.8 & 22.6 & 14.2 & 20.0 & 22.2 & 17.9 & 10.4 & 32.2 & 35.1 & 20.9\\

& NCNet~\cite{Rocco:TPAMI:2020:ncnet} & 
17.9 & 12.2 & 32.1 & 11.7 & 29.0 & 19.9 & 16.1 & 39.2 & 9.9 & 23.9 & 18.8 & 15.7 & 17.4 & 15.9 & 14.8 & 9.6 & 24.2 & 31.1 & 20.1\\

& SFNet~\cite{Lee:CVPR:2019:sfnet} & 
26.9 & 17.2 & 45.5 & 14.7 & \underline{38.0} & 22.2 & 16.4 & \underline{55.3} & 13.5 & 33.4 & 27.5 & 17.7 & 20.8 & 21.1 & 16.6 & 15.6 & 32.2 & 35.9 & 26.3 \\

& PMD~\cite{Li:CVPR:2021:PMD} & 
26.2 & 18.5 & 48.6 & 15.3 & \textbf{38.0} & 21.7 & 17.3 & 51.6 & 13.7 & 34.3 & 25.4 & 18.0 & 20.0 & 24.9 & 15.7 & 16.3 & 31.4 & \underline{38.1} & 26.5\\

& PSCNet-SE~\cite{Jeon:TPAMI:2021:parnj} & 
28.3 & 17.7 & 45.1 & 15.1 & 37.5 & 30.1 & \underline{27.5} & 47.4 & 14.6 & 32.5 & 26.4 & 17.7 & 24.9 & 24.5 & 19.9 & 16.9 & 34.2 & 37.9 & 27.0\\

\hline

\multirow{5}{*}{\parbox{3.2cm}{\centering Weak supervision \\ (test-time optimization)}} &
VGG+MLS~\cite{Aberman:TOG:2018:bestbuddies} & 
29.5 & 22.7 & 61.9 & \textbf{26.5} & 20.6 & 25.4 & 14.1 & 23.7 & 14.2 & 27.6 & 30.0 & 29.1 & 24.7 & 27.4 & 19.1 & 19.3 & 24.4 & 22.6 & 27.4\\

& DINO+MLS~\cite{Aberman:TOG:2018:bestbuddies,Caron:NIPS:2020:unsupervised} & 
49.7 & 20.9 & 63.9 & 19.1 & 32.5 & 27.6 & 22.4 & 48.9 & 14.0 & 36.9 & 39.0 & 30.1 & 21.7 & 41.1 & 17.1 & 18.1 & 35.9 & 21.4 & 31.1\\

& DINO+NN~\cite{Amir:2021:VITdescriptoers} & 
\underline{57.2} & 24.1 & \underline{67.4} & 24.5 & 26.8 & 29.0 & 27.1 & 52.1 & \underline{15.7} & \underline{42.4} & \underline{43.3} & 30.1 & 23.2 & 40.7 & 16.6 & 24.1 & 31.0 & 24.9 & 33.3\\

& NeuCongeal~\cite{Ofri:CVPR:2023:neuracongealingl} & 
- & \textbf{29.1$^\star$} & - & - & - & - & - & 53.3 & - & - & 35.2 & - & - & - & - & - & - & - & -\\

& ASIC~\cite{Gupta:ICCV:2023:ASIC} & 
\textbf{57.9} & \underline{25.2} & \textbf{68.1} & \underline{24.7} & 35.4 & 28.4 & \textbf{30.9} & 54.8 & \textbf{21.6} & \textbf{45.0} & \textbf{47.2} & \underline{39.9} & \underline{26.2} & \textbf{48.8} & 14.5 & \underline{24.5} & \underline{49.0} & 24.6 & \underline{36.9}\\
& SpaceJAM (DINOv1, ViT-S) & 
{43.4} & 
{22.0} & 
{34.5} & 
{18.8} & 
{32.4} & 
\textbf{36.3} & 
{24.1} & 
\textbf{60.8} & 
{06.5} & 
{37.1} & 
{35.9} & 
\textbf{46.7} & 
\textbf{53.6} & 
\underline{46.9} & 
\textbf{36.0} & 
\textbf{24.6} & 
\textbf{75.0} & 
\textbf{46.0} &  
\textbf{37.8} \\ 

\cline{2-21}

& SpaceJAM (DINOv2, ViT-B) &
{52.9} & 
{48.9} & 
{48.1} & 
{25.1} & 
{45.8} & 
{51.8} & 
{42.6} & 
{66.9} & 
{29.8} & 
{51.8} & 
{33.7} & 
{32.6} & 
{57.7} & 
{31.8} & 
{36.9} & 
{10.2} & 
{73.3} & 
{15.7} & 
{42.0} \\ 

 & SpaceJAM (DINOv2, ViT-L) &
{53.6} & 
{53.4} & 
{45.4} & 
{47.5} & 
{71.0} & 
{54.0} & 
{46.0} & 
{66.0} & 
{25.8} & 
{48.6} & 
{28.5} & 
{47.6} & 
{54.0} & 
{50.7} & 
{34.0} & 
{9.0} & 
{71.8} & 
{15.4} & 
{45.7} \\ 

\bottomrule
\end{tabular}}
\end{table}
\subsection{Complexity}\label{results:subsec:complexity}
SpaceJAM is a lightweight model, consisting of only 16K (\ie, 0.016M) trainable parameters.
We compare our method to~\cite{Ofri:CVPR:2023:neuracongealingl,Gupta:ICCV:2023:ASIC} using the official implementations. 
As~\autoref{tab:comparison_methods} shows, SpaceJAM's \# of trainable parameters is \emph{much} lower than 
those of~\cite{Ofri:CVPR:2023:neuracongealingl} and~\cite{Gupta:ICCV:2023:ASIC} which are 28.7M and  7.9M, respectively. \autoref{tab:comparison_methods} also shows that SpaceJAM achieves convergence within only 300 AE pretraining epochs and 400 AE+STN epochs, translating to a training time of $\approx 6$ minutes on a single RTX4090, which starkly contrasts with the \emph{hours} required by the competing methods. Despite its compact design, SpaceJAM attains a PCK@0.1 score of 60.8 on the `Cats' dataset, outperforming both competitors. Evidently, SpaceJAM's efficiency does not hurt performance, as it not only accelerates the training process but also enhances alignment accuracy.

\subsection{Ablation Study \& Limitations}
\textbf{Ablation.} We conducted an ablation study for SpaceJAM (using DINOv2~\cite{Oquab:2023:dinov2}) on subsets of the CUB-200 and SPair-71k datasets and evaluated them using PCK@0.1, as shown in ~\autoref{tab:ablation}.
Due to the low-capacity of the models used in SpaceJAM, the AE pre-training proves essential before the JA. Predicting a cascade of 5 small transformations shows significant improvement over a single large one (STN) or three (ICSTN-3), when the \# of trainable parameters is the same. An alignment to an atlas, even when using the robust Inverse Consistency Averaging Error (ICAE~\cite{Shapira:ICML:2023:RFDTAN}) performs worse than our inverse-compositional framework. \autoref{tab:ablation} also shows that using the Lie groups is crucial and that the curriculum learning's effect, while positive, is less significant.
 \def\ASIC{Gupta:ICCV:2023:ASIC}
\def\DINOMLS{Caron:NIPS:2020:unsupervised, Aberman:TOG:2018:bestbuddies}
\def\DINONN{Amir:2021:VITdescriptoers}
\def\VGGMLS{Aberman:TOG:2018:bestbuddies}

\begin{table}[t]
\centering
\resizebox{0.95\textwidth}{!}{
\begin{minipage}[t]{0.45\textwidth}
    \centering
    \tiny
    \caption{A comparison on CUB-200.}
    \label{tab:CUB}
    \begin{tabular}{lc|cc}
    \toprule
    \textbf{Method} & CUB-200 & \textbf{Method} & CUB-200 \\
                    & (3 cate.) &               & (Subsets) \\
    \midrule
    VGG+MLS~\cite{\VGGMLS}   & 25.8 & - & - \\
    DINO+MLS~\cite{\DINOMLS} & 67.0 & -  & - \\
    DINO+NN~\cite{\DINONN}   & 68.3 & GANgealing~\cite{Peebles:CVPR:2022:gangealing} &  56.8 \\
    ASIC~\cite{\ASIC}        & \textbf{75.9} & NeuCongeal~\cite{Ofri:CVPR:2023:neuracongealingl} & \underline{63.6} \\
    SpaceJAM       & \underline{69.6} & SpaceJAM & \textbf{69.9} \\
    \bottomrule
    \end{tabular}
\end{minipage}%
\begin{minipage}[t]{0.45\textwidth}
    \centering
    \tiny
    \caption{Ablation Study.}
    \label{tab:ablation}
    \begin{tabular}{lcc}
    \toprule
    \textbf{Ablation} & CUB-200 & SPair-71k \\
                      & (3 cate.) & (3 cate.) \\
    \midrule
    Alignment to atlas & 58.5      & 54.6 \\
    No AE pre-training & 54.9      & 56.6 \\
    No Curiculum & 70.8   & 57.2 \\
    No Lie Groups      & 65.6     & 57.1 \\
    STN                & 61.6      & 43.5 \\
    ICSTN-3            & 72.6      & 57.6 \\
    Complete model     & \textbf{73.3} & \textbf{58.1} \\
    \bottomrule
    \end{tabular}
\end{minipage}%
}
\end{table}

{\textbf{Limitations.} }
SpaceJAM, while being an JA effective method, was not designed for pairwise dense correspondences. Thus, it was not optimized for maximizing the per-pixel agreement between an image pair. 
That said, \autoref{Sec:Results} shows
that the speed gains in speed are drastic and that the method is much faster than contemporary JA methods~\cite{Ofri:CVPR:2023:neuracongealingl,Gupta:ICCV:2023:ASIC} while its dense mapping score (a standard evaluation metric for both pairwise and JA tasks) is still comparable (and in fact, in some cases better). This trade-off also comes into play in the choice of the transformation family. As with other STNs~\cite{Jaderberg:NIPS:2015:STN,Schwobel:2022:UAI:pstn,Skafte:CVPR:2018:DDTN}, it is possible to incorporate more expressive transformations such as diffeomorphisms (as in, \eg,~\cite{Freifeld:ICCV:2015:CPAB,Freifeld:PAMI:2017:CPAB,Skafte:CVPR:2018:DDTN}) which are richer than homographies. Lastly, our performance also partially relies on the quality of the initial masks,
a limitation we share with~\cite{Ofri:CVPR:2023:neuracongealingl,Gupta:ICCV:2023:ASIC}.

\section{Conclusion}\label{Sec:Conclusion}
SpaceJAM is a lightweight method that effectively performs JA with neither regularization terms nor building an explicit atlas during the process. It demonstrates excellent performance on benchmarks like SPair-71K and CUB, on par with, or outperforming, existing methods despite having far less trainable parameters and a much shorter running time. 
While here we mentioned only 3 matrix groups, both our method and code support 
additional ones (\eg, the similarity group).

\newpage
\subsubsection{Acknowledgements.} This work was supported by the Lynn and William Frankel
Center at BGU CS, by the Israeli Council for Higher Education via the BGU Data Science Research Center, and by
Israel Science Foundation Personal Grant \#360/21. R.S.W.~and S.E.F.~were also funded in part by the BGU Kreitman School Negev
Scholarship.

\clearpage

\title{SpaceJAM: a Lightweight and Regularization-free Method for Fast Joint Alignment of Images \\ Supplemental Material} 

\titlerunning{SpaceJAM: A Method for Fast Joint Alignment of Images}

\author{
Nir Barel* \orcidlink{0000-0003-4751-5653} \and
Ron Shapira Weber* \orcidlink{0000-0003-4579-0678} \and
Nir Mualem\orcidlink{0000-0005-1216-6774} \and
Shahaf E. Finder\orcidlink{0000-0003-0254-1380} \and
Oren Freifeld\orcidlink{0000-0001-9816-9709}
}

\authorrunning{N.~Barel \etal}

\institute{The Department of Computer Science, Ben-Gurion University of the Negev, Israel\\
\email{\{banir,ronsha,nirmu,finders\}@post.bgu.ac.il, orenfr@cs.bgu.ac.il}}

\maketitle

\section{Handling flips}\label{subsec:implement:flips}
As detailed in the main manuscript, flips require special care. 
Let $(I_i, I_j)$ be a source and target images respectively (for simplicity, we drop the DINO ViT and learned features notion
and use this notion). In this particular case, our loss function reduces to
\begin{align}\label{eq:loss:full:supmat}
   \Lcal_\mathrm{IC} =\ellTwoNorm{I_j - (I_i \circ T^{\btheta_i}
   \circ
   T^{-\btheta{_j}})  
   }^2 \, .
\end{align}
To incorporate flips efficiently, we consider only horizontal flips (since vertical flips could be reached through a horizontal flip + rotation) and compute the gradient only between the best matching pair. Particularly, let $F^{k_{i}}$ be the $k^{\mathrm{th}}$ flip configuration applied to the $i^{\mathrm{th}}$ image, where $k\in C$ such that $C$ holds the possible configuration (in our case, 2). 
The objective function is now 

\begin{align}\label{eq:loss:flips:supmat}
   \Lcal_\mathrm{IC} =\sum_{i,j}\sum_{k_i=1}^{|C|}\min_{k_j \in C } \ellTwoNorm{I_j\circ F^{k_j} - ((I_i \circ T^{\btheta_{i,k_i}}
   )\circ F^{k_i})\circ T^{-\btheta_{j,k_j}} 
   }^2 \, ,
\end{align}

where $(k_i, k_j)$ are the flips considered for the image pair ($I_i, I_j$). 

\section{Curriculum learning}\label{subsec:implement:curriculum}
\begin{figure}[h]
\centering
\def\figwidth{0.99\linewidth}
    \includegraphics[trim = 0mm 0mm 0mm 0mm, clip, width=\figwidth]{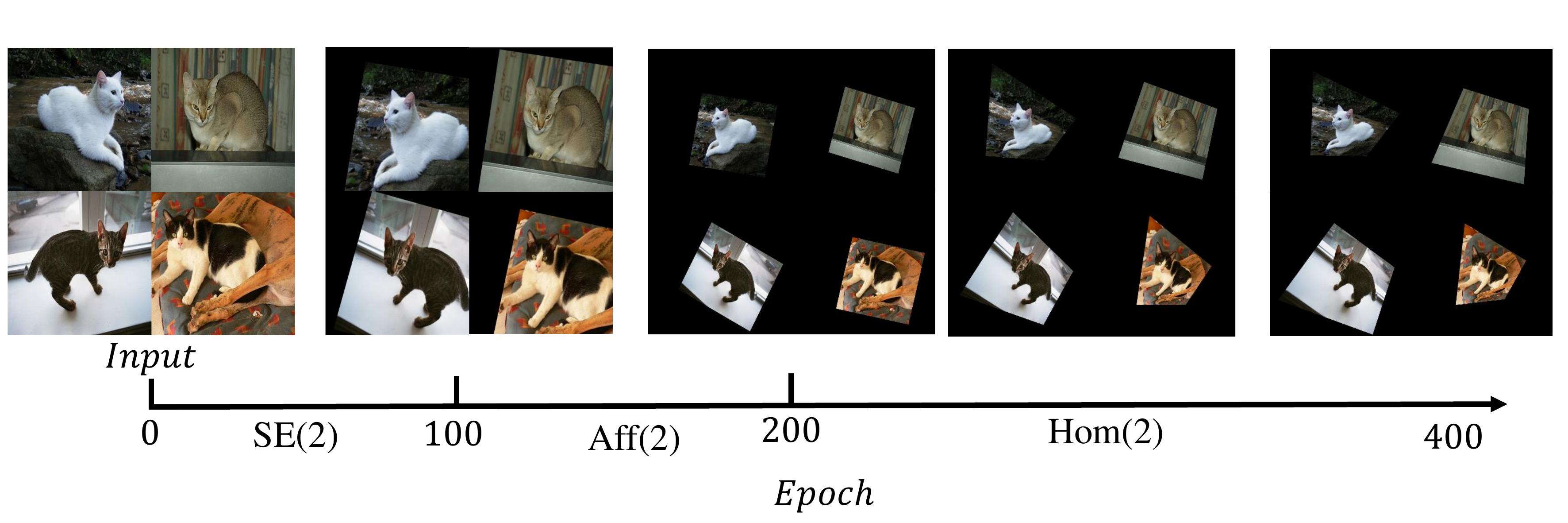}

    \caption{
    Lie Algebric curriculum learning. The notation $\mathrm{SE}(2)$ between the epochs $(0,100)$ states that during that interval, the training is restricted to $\mathrm{SE}(2)$. At epoch 100, more transformation parameters are ``released" to allow for affine transformations.
    }
\label{fig:training:curriculum}
\end{figure}

To incorporate the Lie-algebraic curriculum learning during training, we gradually add more complex transformation modules, starting from $\mathrm{SE}(2)$ and later ``release" more of the transformation parameters, to obtain (invertible) affine transformations and finally homographies.~\autoref{fig:training:curriculum} illustrate the process, where additional transformation parameters are ``released", as illustrated by the warped images above the training timeline. 

\section{Inverse-Compositional STN (IC-STN)~\cite{Lin:CPVR:2017:inverse}}
The IC-STN~\cite{Lin:CPVR:2017:inverse}, based on the classical Inverse-Compositional Lucas \& Kanade algorithm~\cite{Lucas:IJCAI:1981:iterative}, predicts a cascade of smaller warps and composes them. 
In our case, we do so by recursively feeding the STN several times with its own output. 
In effect, 
\begin{align}
(\btheta, X \circ T^{\btheta})&=\psi_{\mathrm{STN}}(X)\\
(\btheta', X \circ T^{\btheta} \circ T^{\btheta'})&=\psi_{\mathrm{STN}}(X \circ T^{\btheta} ) \\
(\btheta'', X\circ T^{\btheta}\circ T^{\btheta'}\circ T^{\btheta''})&=\psi_{\mathrm{STN}}(X \circ T^{\btheta}\circ T^{\btheta'})
\end{align}
and so on.

\section{An Additional runtime comparison}
We provide an additional runtime comparison with Neural Congealing~\cite{Ofri:CVPR:2023:neuracongealingl} and ASIC~\cite{Gupta:ICCV:2023:ASIC} on the `Dog'  and `Bike' datasets~\cite{Min:2019:spair}. The results are presented in~\autoref{submat:tab:comparison_methods}.
\begin{table}[h]
\newcommand{\cmark}{\ding{51}}%
\newcommand{\xmark}{\ding{55}}%
\centering
\caption{A comparison with recent JA methods and evaluation on 3 SPair-71K categories~\cite{Min:2019:spair}.}
\scriptsize
\label{submat:tab:comparison_methods}
\begin{tabular}{lcccccccc}
\toprule
\textbf{Method} & \# Params & \# Losses & \#HP & Atlas-free learning & \#epochs & \multicolumn{3}{|c|}{Time} \\ 
& & & & & & Cat &Bike & Dog \\
\midrule
 NeuCongealing~\cite{Ofri:CVPR:2023:neuracongealingl} &$28.7\mathrm{M}$ & 8 & 8 & \xmark  & 8K & 1:17:02  & 1:12:55 &  1:25:28 \\ 
 \midrule
 ASIC~\cite{Gupta:ICCV:2023:ASIC} &  $7.9\mathrm{M}$& 4 & 5 & \xmark  & 20K  & 1:06:48 & 1:07:40 & 1:06:11 \\ 
 \midrule
 SpaceJAM (Ours) & \textbf{0.016M} &  \textbf{1} &  \textbf{0} & \cmark  & \textbf{0.7K} & \textbf{0:05:58}  &  \textbf{0:06:11} & \textbf{00:05:43} \\ 
 \bottomrule
\end{tabular}
\end{table}

\section{Architectures}\label{subsec:implement:arch}
A detailed overview of the Autoencoder (AE) and Spatial Transformer Network (STN) architectures and the number of trainable parameters. Together they form our alignment module - $\psi_{\mathrm{align}}$. 
\begin{table}[ht!]
\centering
\resizebox{0.98\textwidth}{!}{
\begin{minipage}[t]{0.49\linewidth}
\centering
\scriptsize
\caption{Autoencoder Model Summary.\\ GFMN = GlobalFeatureMapNormalizer.}
\begin{tabular}{llr}
\toprule
Layer (type:depth-idx) & Output Shape & Param \\
\midrule
    Autoencoder & [1, 25, 256, 256] & -- \\
    \,\, Encoder: 1-1 & [1, 3, 256, 256] & -- \\
    \,\,\,\, Sequential: 2-1 & [1, 3, 256, 256] & -- \\
    \,\,\,\,\,\, Conv2d: 3-1 & [1, 32, 256, 256] & 832 \\
    \,\,\,\,\,\, ReLU: 3-2 & [1, 32, 256, 256] & -- \\
    \,\,\,\,\,\, BatchNorm2d: 3-3 & [1, 32, 256, 256] & 64 \\
    \,\,\,\,\,\, Conv2d: 3-4 & [1, 16, 256, 256] & 528 \\
    \,\,\,\,\,\, ReLU: 3-5 & [1, 16, 256, 256] & -- \\
    \,\,\,\,\,\, BatchNorm2d: 3-6 & [1, 16, 256, 256] & 32 \\
    \,\,\,\,\,\, Conv2d: 3-7 & [1, 3, 256, 256] & 51 \\
    \,\,\,\,\,\, GFMN: 3-8 & [1, 3, 256, 256] & -- \\
    \,\, Decoder: 1-2 & [1, 25, 256, 256] & -- \\
    \,\,\,\, Sequential: 2-2 & [1, 25, 256, 256] & -- \\
    \,\,\,\,\,\, Conv2d: 3-9 & [1, 16, 256, 256] & 64 \\
    \,\,\,\,\,\, ReLU: 3-10 & [1, 16, 256, 256] & -- \\
    \,\,\,\,\,\, BatchNorm2d: 3-11 & [1, 16, 256, 256] & 32 \\
    \,\,\,\,\,\, Conv2d: 3-12 & [1, 32, 256, 256] & 544 \\
    \,\,\,\,\,\, ReLU: 3-13 & [1, 32, 256, 256] & -- \\
    \,\,\,\,\,\, BatchNorm2d: 3-14 & [1, 32, 256, 256] & 64 \\
    \,\,\,\,\,\, Conv2d: 3-15 & [1, 25, 256, 256] & 825 \\
\bottomrule
\end{tabular}
\end{minipage}%
\hfill%
\begin{minipage}[t]{0.49\linewidth}
    \centering
    \scriptsize
    \caption{STN Model Summary.}
    \label{tab:archs}
    \begin{tabular}{llr}
    \toprule
    Layer (type:depth-idx) & Output Shape & Param \\
    \midrule
    Conv2d-1 & [1, 10, 250, 250] & 1,480 \\
    AdaptiveMaxPool2d-2 & [1, 10, 32, 32] & 0 \\
    ReLU-3 & [1, 10, 32, 32] & 0 \\
    Conv2d-4 & [1, 5, 28, 28] & 1,255 \\
    AdaptiveMaxPool2d-5 & [1, 5, 8, 8] & 0 \\
    ReLU-6 & [1, 5, 8, 8] & 0 \\
    \hline
    Linear-1 & [1, 1, 32] & 10,272 \\
    ReLU-2 & [1, 1, 32] & 0 \\
    Linear-3 & [1, 1, 9] & 297 \\
    \hline
    \end{tabular}
\end{minipage}%
}
\end{table}

\begin{figure}[h]
\centering
\def\figwidth{0.9\linewidth}
    \includegraphics[trim = 0mm 0mm 0mm 0mm, clip, width=\figwidth]{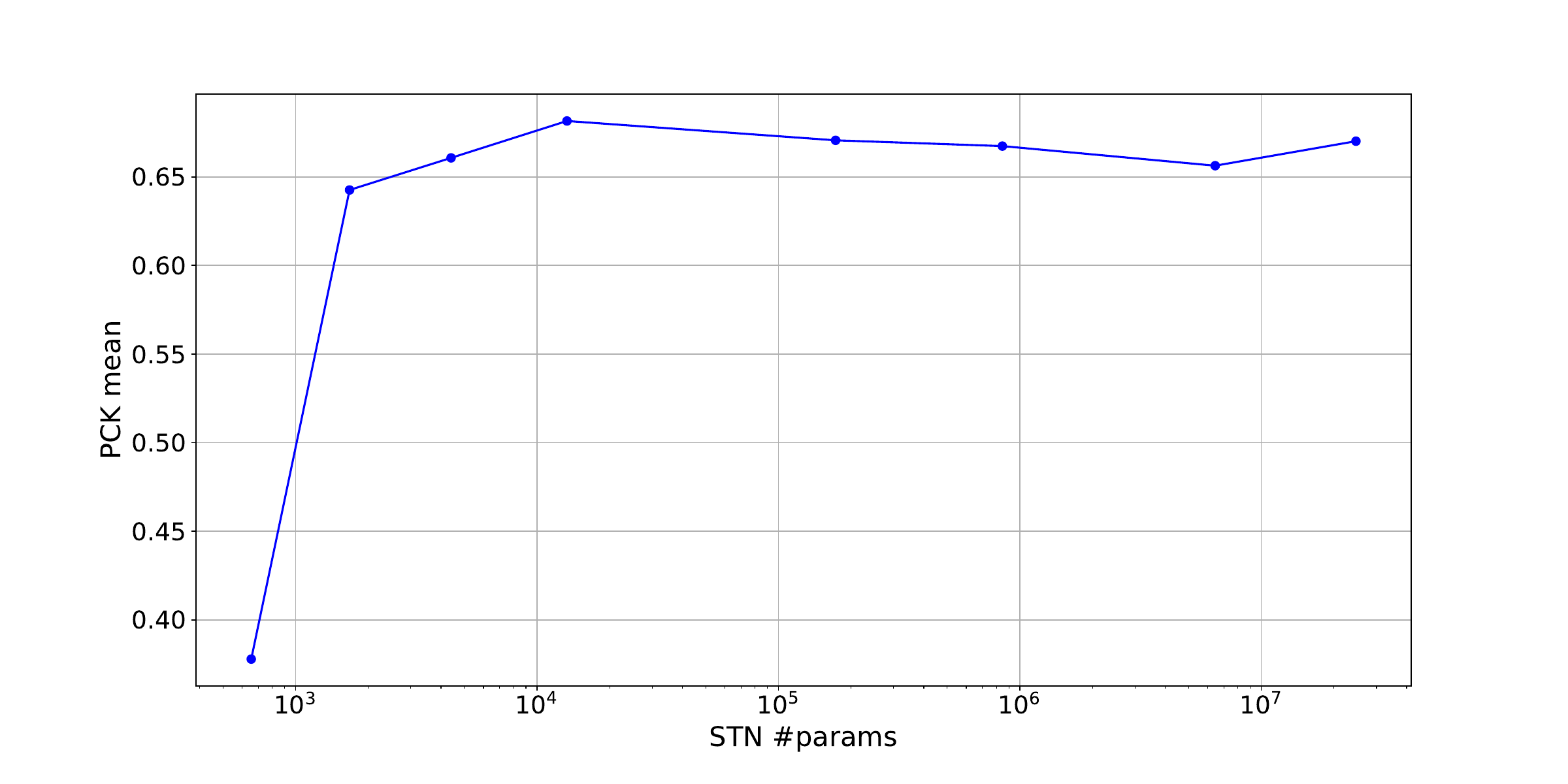}

    \caption{
    Average PCK@0.1 score as a function of  \# trainable parameters of the STN (the x-axis is log-scaled). The model reaches saturation around $\sim 15$K parameters. 
    }
\label{fig:pck:stn:size}
\end{figure}

We also evaluate the effect of the STN size on the resulting alignment.~\autoref{fig:pck:stn:size} shows the average PCK@0.1 of 5 runs for the 3 subsets of the CUB200 datasets~\cite{Wah:2011:CUB}. We increase the number of trainable parameters by using the same STN architecture with larger convolutional blocks in terms of \# kernels and their size. Notably, the performance effectively saturates at as early as $\sim$15K parameters. Increasing the model further even to 24M parameters, does not results in additional gains.  
\section{Further Discussion of the Results}
\begin{figure}[t]
    \centering
    \begin{subfigure}{0.3\linewidth}
        \centering
        \includegraphics[width=\linewidth]{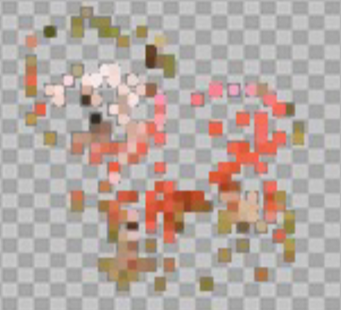}
        \caption{DINO+NN}
        \label{fig:image1}
    \end{subfigure}
    \hfill
    \begin{subfigure}{0.3\linewidth}
        \centering        \includegraphics[width=\linewidth]{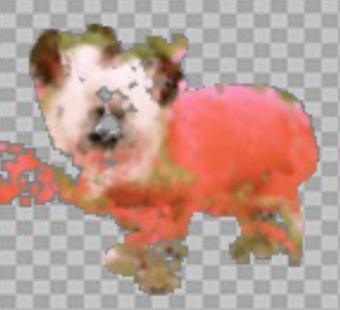}
        \caption{ASIC}
        \label{fig:image2}
    \end{subfigure}
    \hfill
    \begin{subfigure}{0.3\linewidth}
        \centering        \includegraphics[width=\linewidth]{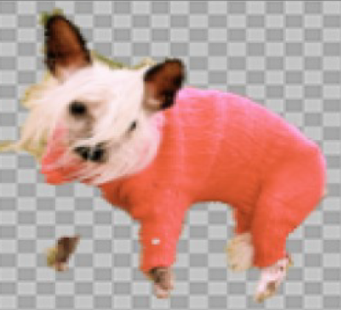}
        \caption{SpaceJAM (ours)}
        \label{fig:image3}
    \end{subfigure}
    \caption{Geometric fidelity of transformed images (DINO+NN and ASIC results were obtained from~\cite{Gupta:ICCV:2023:ASIC}).}
    \label{fig:geometric:fidelity}
\end{figure}

A natural question arises -- why do different models perform better in some classes and worse in others? For instance, consider the 'Dogs' class of the SPair dataset.
In the results presented in Table 2 in the main paper, ASIC outperforms the proposed method on that class by approximately 11 points, suggesting superior alignment. However, the visual comparison in Figure 3 reveals that dense-correspondence methods like ASIC often result in incoherent alignment. 
The warped images display artifacts such as holes, and the dog faces become unrecognizable. This discrepancy arises because benchmarks like SPair and CUB-200 focus on the sparse correspondence of hand-picked key points rather than measuring global alignment. 
In fact, the basic DINO-NN outperforms SpaceJAM on the same `Dog' class, but yields significantly poorer visual results, as illustrated in~\autoref{fig:geometric:fidelity}. This highlights the limitations of the DINO-NN approach. Additionally, our method outperforms ASIC in more than half of the classes while requiring 100x fewer parameters and achieving a 10x reduction in training time. 
Finally, the variance in results can also be attributed to the small set size (20-30 images) compared to the diverse poses, illuminations, and occlusions present in each set.

\clearpage

\section{Additional Visualizations}\label{Sec:visual}
\subsection{Additional joint alignment results}\label{supmat:subsec:ja:figs}
More visual results of SpaceJAM's joint alignment (JA) are presented below (Figures 2-6) for the SPair-71K~\cite{Min:2019:spair} and Samurai (`robot')~\cite{Boss:NIPS:2022:samurai} datasets.
The figures show, from top-to-bottom: 1) input images; 2) DINO ViT features (first 3 PCs); 3) learned features 4) aligned features, and  5) aligned images. The aligned features and images are masked by the intersection of the coarse input mask and the median mask of the set (both after alignment). The atlas of the set appears at the bottom right. 

\def\parentdirs{"train", "cat", "robot", "plane", "bus"}


\foreach \parentdir in \parentdirs{
    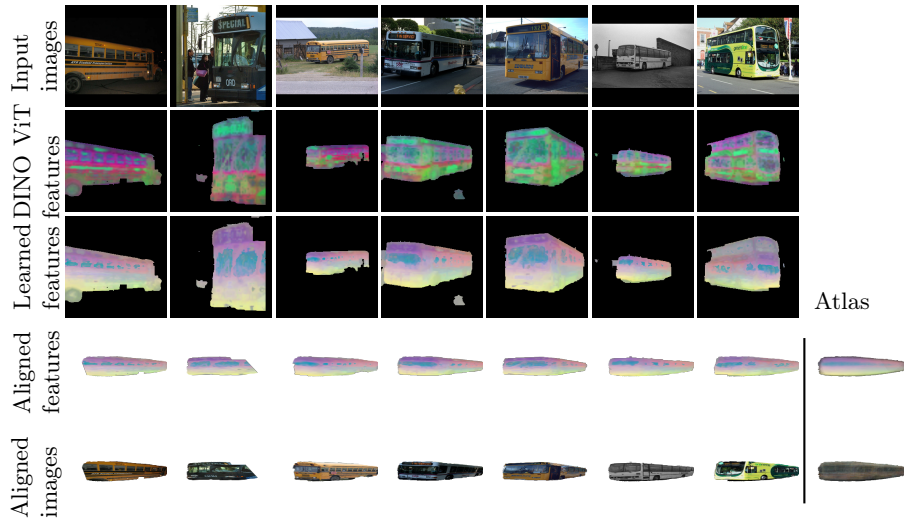
\begin{figure}[ht!]
    \centering

    \begin{tikzpicture}
    \def\numrows{5}
    
    \def\numcols{8}

    \def\rowcaptions{{"Input\\images", "DINO ViT \\ features", "Learned\\features", "Aligned\\features", "Aligned\\images", "Encoded\\overlay"}}
    \def\directories{{"imgs", "dino_keys", "encoded", "encoded_warped", "img_warped"}}
    \def\imagenames{{"00", "01", "02", "03","04","05","06","99","07","09", "10","11"}}

    \foreach \row in {1,...,\numrows}{
        \pgfmathsetmacro\ypos{-(\row - 1) * 1.4 - 0.75}
        
        \node[rotate=90, align=left] at (-0.41, \ypos) {\pgfmathparse{\rowcaptions[\row-1]}\pgfmathresult};

        \pgfmathparse{\directories[\row-1]}
        \edef\currentdir{\pgfmathresult}
        \xdef\currentdir{./figures_supmat/\parentdir/\currentdir} 
        
        \foreach \col in {1,...,\numcols}{
            \pgfmathsetmacro\xpos{\col * 1.4 - 0.75}
            
            \pgfmathsetmacro\imagename{\imagenames[\col-1]}

            \node[inner sep=0pt] (image-\row-\col) at (\xpos, \ypos) {
                \expandafter\includegraphics\expandafter[width=.11\textwidth]{\currentdir/fig_\imagename.png}
            };
        }
    }
        \node[font=\footnotesize] (atlas-text) at (10.3,-4) {Atlas};
        \draw[thick] (9.8,-6.7) -- (9.8,-4.5);
    \end{tikzpicture}
    \caption{\textbf{Joint alignment results - \parentdir}.}
    \edef\tempLabel{fig:JA-\parentdir}
    \expandafter\label{\tempLabel} 
\end{figure}
}

\clearpage

\subsection{Additional pairwise alignment results}\label{supmat:subsec:pairwise:figs}
More visual results of SpaceJAM's pairwise alignment are presented below. 
The figures show, from top-to-bottom: 1) input images; 2) learned features overlay; 3-7) Source-to-target pairwise alignment, where the image in the red square is aligned to all other images. 

\def\parentdirs{"train", "bus"}


\foreach \parentdir in \parentdirs{

    \begin{figure}[ht!]
    \centering

    \begin{tikzpicture}
    \def\numrows{7}
    
    \def\numcols{5}

    \def\rowcaptions{{"Input\\images", "Learned \\ features", "Pairwise\\alignment", "", "", "", ""}}
    \def\directories{{"imgs", "overlay", "row0", "row1", "row2", "row3", "row4", "row5"}}
    \def\imagenames{{"02","03","04","05","06","07"}}

    \foreach \row in {1,...,\numrows}{
        \pgfmathsetmacro\ypos{-(\row - 1) * 2.1 - 0.75}
        
        \node[rotate=90, align=left] at (-0.41, \ypos) {\pgfmathparse{\rowcaptions[\row-1]}\pgfmathresult};

        \pgfmathparse{\directories[\row-1]}
        \edef\currentdir{\pgfmathresult}
        \xdef\currentdir{./figures_supmat/\parentdir/\currentdir} 
        
        \foreach \col in {1,...,\numcols}{
            \pgfmathsetmacro\xpos{\col * 2.1 - 0.75}
            
            \pgfmathsetmacro\imagename{\imagenames[\col-1]}

            \node[inner sep=0pt] (image-\row-\col) at (\xpos, \ypos) {
                \expandafter\includegraphics\expandafter[width=.17\textwidth]{\currentdir/fig_\imagename.png}
            };
        }
    }

    \node[draw=red, 
          thick, 
          minimum size=2.05cm, 
          anchor=south,
          ] (redsquare) at (1.35,-5.97) {}; 
    \node[draw=red, 
          thick, 
          minimum size=2.05cm, 
          anchor=south,
          ] (redsquare2) at (3.45,-8.07) {}; 
    \node[draw=red, 
          thick, 
          minimum size=2.05cm, 
          anchor=south,
          ] (redsquare3) at (5.55,-10.17) {}; 
    \node[draw=red, 
          thick, 
          minimum size=2.05cm, 
          anchor=south,
          ] (redsquare4) at (7.65,-12.27) {}; 
    \node[draw=red, 
          thick, 
          minimum size=2.05cm, 
          anchor=south,
          ] (redsquare5) at (9.75,-14.37) {}; 

    \node[below=-0.05cm of redsquare, text=red, font=\scriptsize] {Source}; 
    
    \end{tikzpicture}
    \caption{\textbf{Pairwise alignment results - \parentdir}.}
    \edef\tempLabel{fig:pairwise-\parentdir}
    \expandafter\label{\tempLabel}
\end{figure}
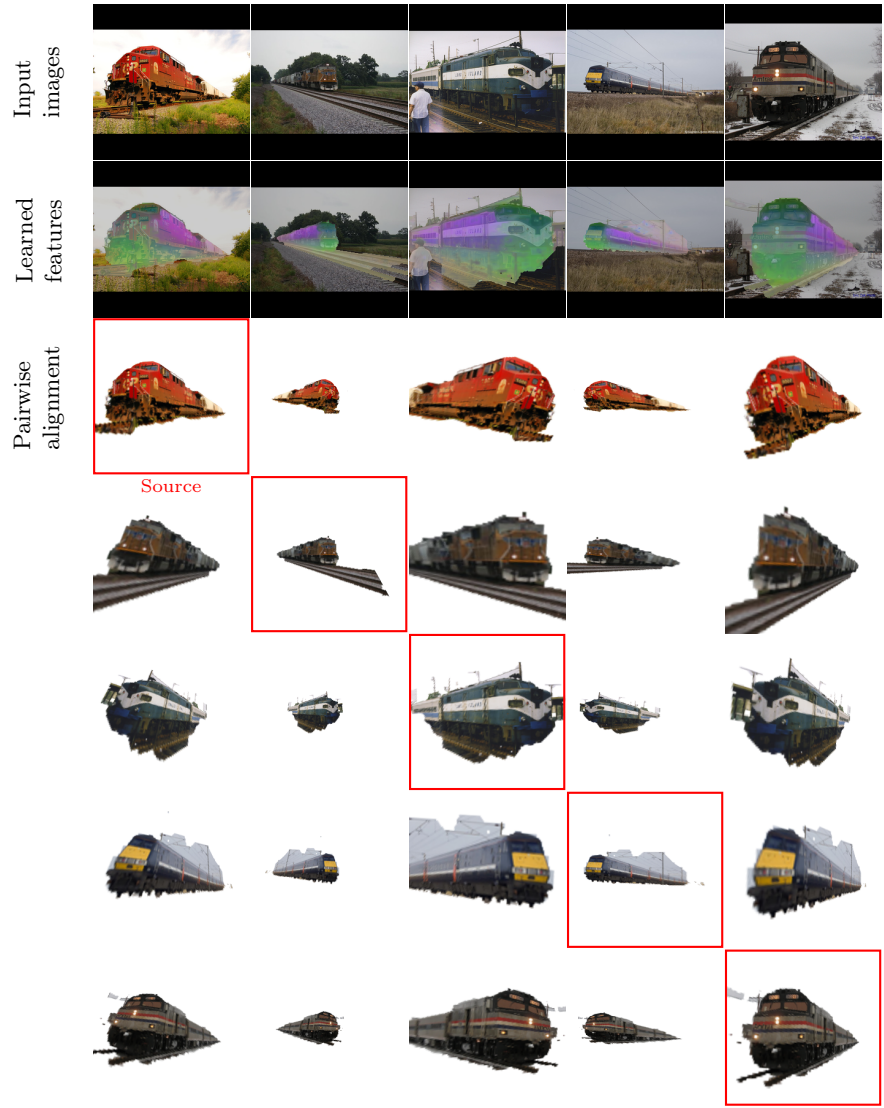
}

\bibliographystyle{splncs04}
\bibliography{refs}
\end{document}